\documentclass[conference]{IEEEtran}
\usepackage{times}
\usepackage[numbers]{natbib}
\usepackage{multicol}
\usepackage[bookmarks=true]{hyperref}
\IEEEoverridecommandlockouts 

\usepackage{url}
\usepackage{amsmath,amssymb}
\usepackage{graphicx,color}
\usepackage{verbatim}
\usepackage{gensymb}
\usepackage{setspace}
\graphicspath{{figures/}}
\newcommand{\ba}{\mathbf{a}}

\newcommand{\bo}{\mathbf{o}}

\newcommand{\bz}{\mathbf{z}}

\usepackage{xspace}
\newcommand{\vsf}{VisuoSpatial Foresight\xspace} 
\newcommand{\il}{Imitation Learning\xspace}

\usepackage{caption}
\def\tablename{Table}

\title{
VisuoSpatial Foresight for Multi-Step,\\ Multi-Task Fabric Manipulation
}

\author{Ryan Hoque$^{1,*}$, Daniel Seita$^{1,*}$, Ashwin Balakrishna$^1$, Aditya Ganapathi$^1$, \\ Ajay Kumar Tanwani$^1$, Nawid Jamali$^2$, Katsu Yamane$^2$, Soshi Iba$^2$, Ken Goldberg$^{1}$
\thanks{$^*$equal contribution.}
\thanks{$^{1}$AUTOLAB at the University of California, Berkeley, USA.}%
\thanks{$^{2}$Honda Research Institute USA, Inc.}%
\thanks{Correspondence to {\tt\small \{ryanhoque,seita\}@berkeley.edu}}%
}

\begin{document}

\maketitle
\thispagestyle{empty}
\pagestyle{empty}

\begin{abstract}
Robotic fabric manipulation has applications in home robotics, textiles, senior care and surgery. Existing fabric manipulation techniques, however, are designed for specific tasks, making it difficult to generalize across different but related tasks. We extend the Visual Foresight framework to learn fabric dynamics that can be efficiently reused to accomplish different fabric manipulation tasks with a single goal-conditioned policy. We introduce VisuoSpatial Foresight (VSF), which builds on prior work by learning visual dynamics on domain randomized RGB images and depth maps simultaneously and completely in simulation. We experimentally evaluate VSF on multi-step fabric smoothing and folding tasks against 5 baseline methods in simulation and on the da Vinci Research Kit (dVRK) surgical robot without any demonstrations at train or test time. Furthermore, we find that leveraging depth significantly improves performance. RGBD data yields an $\mathbf{80 \%}$ improvement in fabric folding success rate over pure RGB data. Code, data, videos, and supplementary material are available at \url{https://sites.google.com/view/fabric-vsf/}.

\end{abstract}

\IEEEpeerreviewmaketitle

\section{Introduction}\label{sec:intro}

Advances in robotic manipulation of deformable objects has lagged behind work on rigid objects due to the far more complex dynamics and configuration space. Fabric manipulation in particular has applications ranging from senior care~\cite{personalized_dressing_2016}, sewing~\cite{sewing_2012}, ironing~\cite{ironing_2016}, bed-making~\cite{seita-bedmaking} and laundry folding~\cite{laundry2012,folding_iros_2015,folding_2017,shibata2012trajectory} to manufacturing upholstery~\cite{Torgerson1987VisionGR} and handling surgical gauze~\cite{thananjeyan2017multilateral}. However, prior work in fabric manipulation has generally focused on designing policies that are only applicable to a \textit{specific} task via manual design \cite{laundry2012,folding_iros_2015,folding_2017,shibata2012trajectory} or policy learning~\cite{seita_ryan, lerrel}. 


The difficulty in developing accurate analytical models of highly deformable objects such as fabric motivates using data-driven strategies to estimate models, which can then be used for general purpose planning. While there has been prior work in system identification for robotic manipulation~\cite{GP-MPC,cautious-MPC,berkenkamp2016safe, MPCRacing, system-id, handful-of-trials}, many of these techniques rely on reliable state estimation from observations, which is especially challenging for deformable objects. One popular alternative is visual foresight~\cite{visual_foresight_2018,finn_vf_2017}, which uses a large dataset of self-supervised interaction data in the environment to learn a visual dynamics model directly from raw image observations and has shown the ability to generalize to a wide variety of environmental conditions~\cite{robonet}. This model can then be used for planning to perform different tasks at test time. The technique has been successfully applied to learning the dynamics of complex tasks, such as pushing and basic fabric folding~\cite{visual_foresight_2018,finn_vf_2017}. However, two limitations of prior work in visual foresight are: 1) the data requirement for learning accurate visual dynamics models is often very high, requiring several days of continuous data collection on real robots~\cite{robonet,visual_foresight_2018}, and 2) while prior work reports experiments on basic fabric manipulation tasks~\cite{visual_foresight_2018}, these are relatively short-horizon tasks with a wide range of valid goal images.

\begin{figure}[t]
\center
\includegraphics[width=0.48\textwidth]{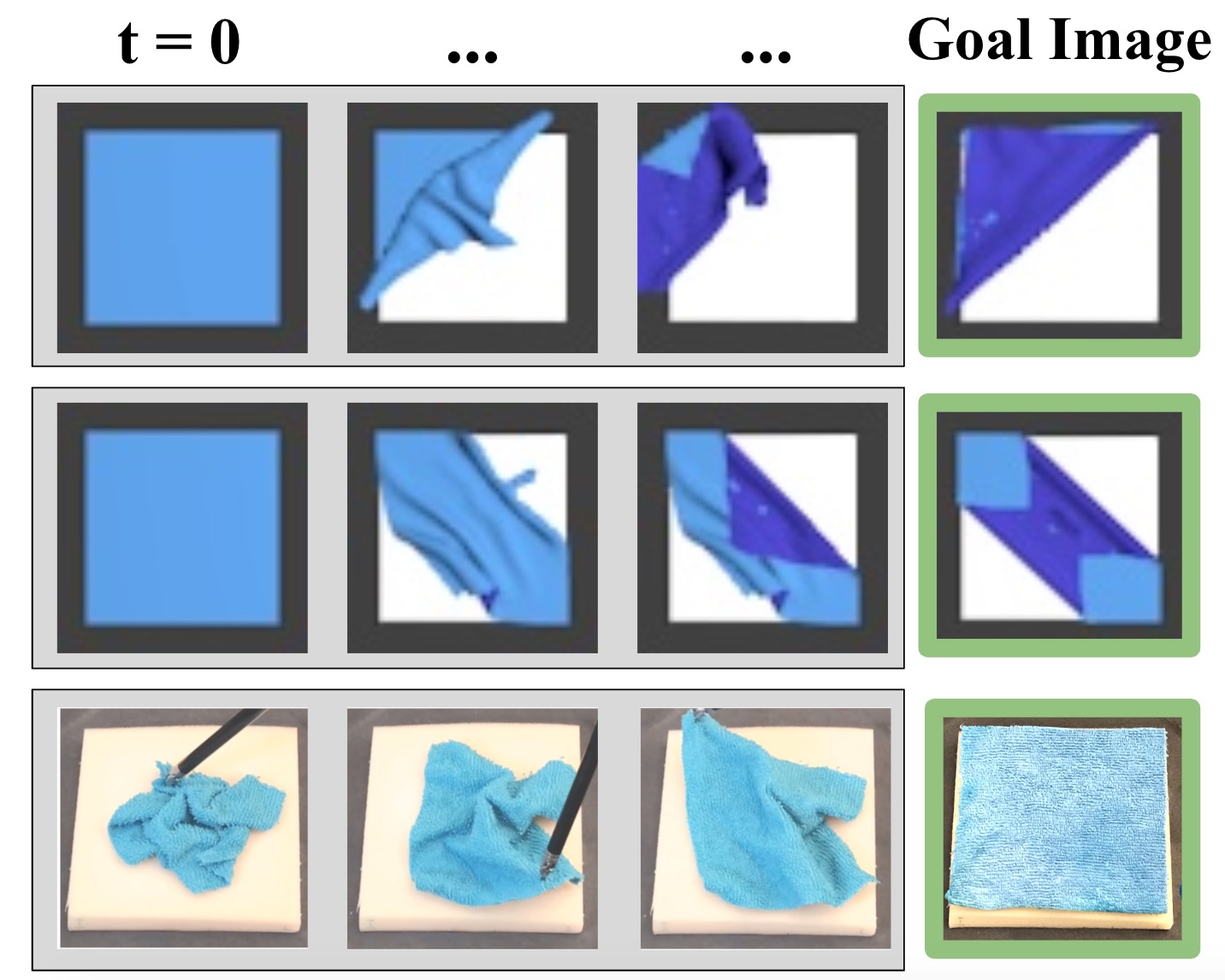}
\caption{
\small
Using VSF on domain randomized RGBD data, we learn a goal-conditioned fabric manipulation policy in simulation without any task demonstrations. We evaluate the same policy on several goal images in simulation (top two rows) and on the physical da Vinci surgical robot (bottom row). The rows display the RGB portions of subsampled frames from episodes for folding, double folding, and smoothing, respectively, toward the goal images in the right column.
}
\vspace*{-10pt}
\label{fig:teaser}
\end{figure}

In this work, we present a system (Figure~\ref{fig:teaser}) that takes steps towards addressing these challenges by integrating RGB and depth sensing to learn visual-spatial (visuospatial) dynamics models in simulation using only random interaction data for training and domain randomization techniques for sim-to-real transfer. This paper introduces: 1) ``\vsf'' on domain randomized, simulated RGB and depth data to facilitate rapid data collection and transfer to physical systems, 2) simulated experiments demonstrating how the learned dynamics can be used to define a goal-conditioned fabric manipulation policy which can perform a variety of multi-step fabric smoothing and folding tasks, and 3) results suggesting that the learned policy transfers to a real physical system with promising smoothing and folding results. In physical smoothing experiments, the learned policy is able to outperform an imitation learning baseline by $2 \%$ despite using no task demonstrations.



\section{Related Work}\label{sec:rw}

Manipulating fabric is a long-standing challenge in robotics. Fabric smoothing in particular has received attention since it helps standardize the configuration of the fabric for subsequent tasks such as folding~\cite{grasp_centered_survey_2019,manip_deformable_survey_2018}. A popular approach in prior work is to first hang fabric in the air and allow gravity to ``vertically smooth'' it~\cite{osawa_2007,kita_2009_iros,kita_2009_icra,unfolding_rf_2014}.~\citet{maitin2010cloth}, use this approach to achieve a 100\% success rate in single-towel folding. In contrast, the approach we present is targeted towards larger fabrics like blankets~\cite{seita-bedmaking} and for single-armed robots which may have a limited range of motion, making such ``vertical smoothing'' infeasible. Similar work on fabric smoothing and folding, such as by~\citet{balaguer2011combining} and Jia~et~al.~\cite{jia_visual_feedback_2018,jia_cloth_manip_2019} assume the robot is initialized with the fabric already grasped, while we initialize the robot's end-effector away from the fabric.~\citet{willimon_unfolding_laundry_2011} and~\citet{heuristic_wrinkles_2014, cloth_icra_2015} address a similar problem setting as we do, but with initial fabric configurations much closer to fully smoothed than those we consider.

There has been recent interest in learning sequential fabric manipulation policies with fabric simulators. For example,~\citet{seita_ryan} and~\citet{lerrel} learn fabric smoothing in simulation, the former using DAgger~\cite{ross2011reduction} and the latter using model-free reinforcement learning (MFRL). Similarly,~\citet{sim2real_deform_2018} and~\citet{rishabh_2019} learn policies for folding fabrics using MFRL augmented with task-specific demonstrations. All of these works obtain large training datasets from fabric simulators; examples of simulators with support for fabric include ARCSim~\cite{arcsim2012}, MuJoCo~\cite{mujoco}, PyBullet~\cite{coumans2019}, and Blender~\cite{blender}. While these algorithms achieve impressive results, they are designed or trained for specific fabric manipulation tasks (such as folding or smoothing), and do not reuse learned structure to generalize to a wide range of tasks. This motivates learning fabric dynamics to enable more general purpose fabric manipulation strategies. 

Model-predictive control (MPC) is a popular approach for leveraging learned dynamics for robotics control that has shown success in learning robust closed-loop policies even with substantial dynamical uncertainty~\cite{thananjeyan2019safety, converging-supervisor, deep_dressing_2018,rosen_icra_tissues_2019}. However, many of these prior works require knowledge or estimation of underlying system state, which can often be infeasible or inaccurate. \citet{finn_vf_2017} and~\citet{visual_foresight_2018} introduce \textit{visual foresight}, and demonstrate that MPC can be used to plan over learned video prediction models to accomplish a variety of robotic tasks, including deformable object manipulation such as folding pants. However, the trajectories shown in \citet{visual_foresight_2018} are limited to a single pick and pull, while we focus on longer horizon sequential tasks that are enabled by a pick-and-pull action space rather than direct end effector control. Furthermore, the fabric manipulation tasks reported have a wide range of valid goal configurations, such as covering a utensil with a towel or moving a pant leg upwards. In contrast, we focus on achieving precise goal configurations via multi-step interaction with the fabric. Prior work on visual foresight~\cite{finn_vf_2017, visual_foresight_2018, robonet} also generally collects data for training visual dynamics models in the real world, which is impractical and unsafe for robots such as the da Vinci surgical robot due to the sheer volume of data required for the technique (on the order of 100,000 to 1 million actions, often requiring several days of physical interaction data~\cite{robonet}). One recent exception is the work of~\citet{time_reversal} which trains visual dynamics models in simulation for Tetris block matching. Finally, prior work in visual foresight utilizes visual dynamics model for RGB images, but we find that depth data provides valuable geometric information for fabric manipulation tasks. 



\section{Problem Statement}\label{sec:bg}

We consider learning goal-conditioned fabric manipulation policies that enable planning to specific fabric configurations given a goal image of the fabric in the desired configuration. We define the fabric configuration at time $t$ as $\boldsymbol{\xi}_t$, represented via a mass-spring system with an $N \times N$ grid of point masses subject to gravity and Hookean spring forces. Due to the difficulties of state estimation for highly deformable objects such as fabric, we consider overhead RGBD observations $\bo_t \in \mathbb{R}^{56 \times 56 \times 4}$, which consist of three-channel RGB and single-channel depth images.

We assume tasks have a finite task horizon $T$ and can be achieved with a sequence of actions (from a single robot arm) which involve grasping a specific point on the fabric and pulling it in a particular direction, which holds for common manipulation tasks such as folding and smoothing. We consider four dimensional actions, 
\begin{equation}
\ba_t = \langle x_t, y_t, \Delta x_t, \Delta y_t\rangle.
\end{equation}
Each action $\ba_t$ at time $t$ involves grasping the top layer of the fabric at coordinate $(x_t, y_t)$ with respect to an underlying background plane, lifting, translating by $(\Delta x_t, \Delta y_t)$, and then releasing and letting the fabric settle. 
When appropriate, we omit the time subscript $t$ for brevity.

The objective is to learn a goal-conditioned policy which minimizes some goal conditioned cost function $c_g(\tau$) defined on realized interaction episodes with goal $g$ and episode $\tau$, where the latter in this work consists of one or more images.

\section{Approach}\label{sec:approach}

\subsection{Goal Conditioned Fabric Manipulation}\label{subsec:goal-cond}
To learn goal-conditioned policies, we build on the visual foresight framework introduced by~\citet{finn_vf_2017}. Here, a video prediction model (also called a visual dynamics model) is trained, which, given a history of observed images and a sequence of proposed actions, generates a sequence of predicted images that would result from executing the proposed actions in the environment. This model is learned from random interaction data of the robot in the environment. Then, MPC is used to plan over this visual dynamics model with some cost function evaluating the discrepancy between predicted images and a desired goal image. In this work, we: 1) learn a visual dynamics model on RGBD images instead of RGB images as in prior work, and 2) learn visual dynamics entirely in simulation. We find that these choices improve performance on complex fabric manipulation tasks and accelerate data collection while limiting wear and tear on the physical system.

\begin{figure}[t]
\center
\includegraphics[width=0.48\textwidth]{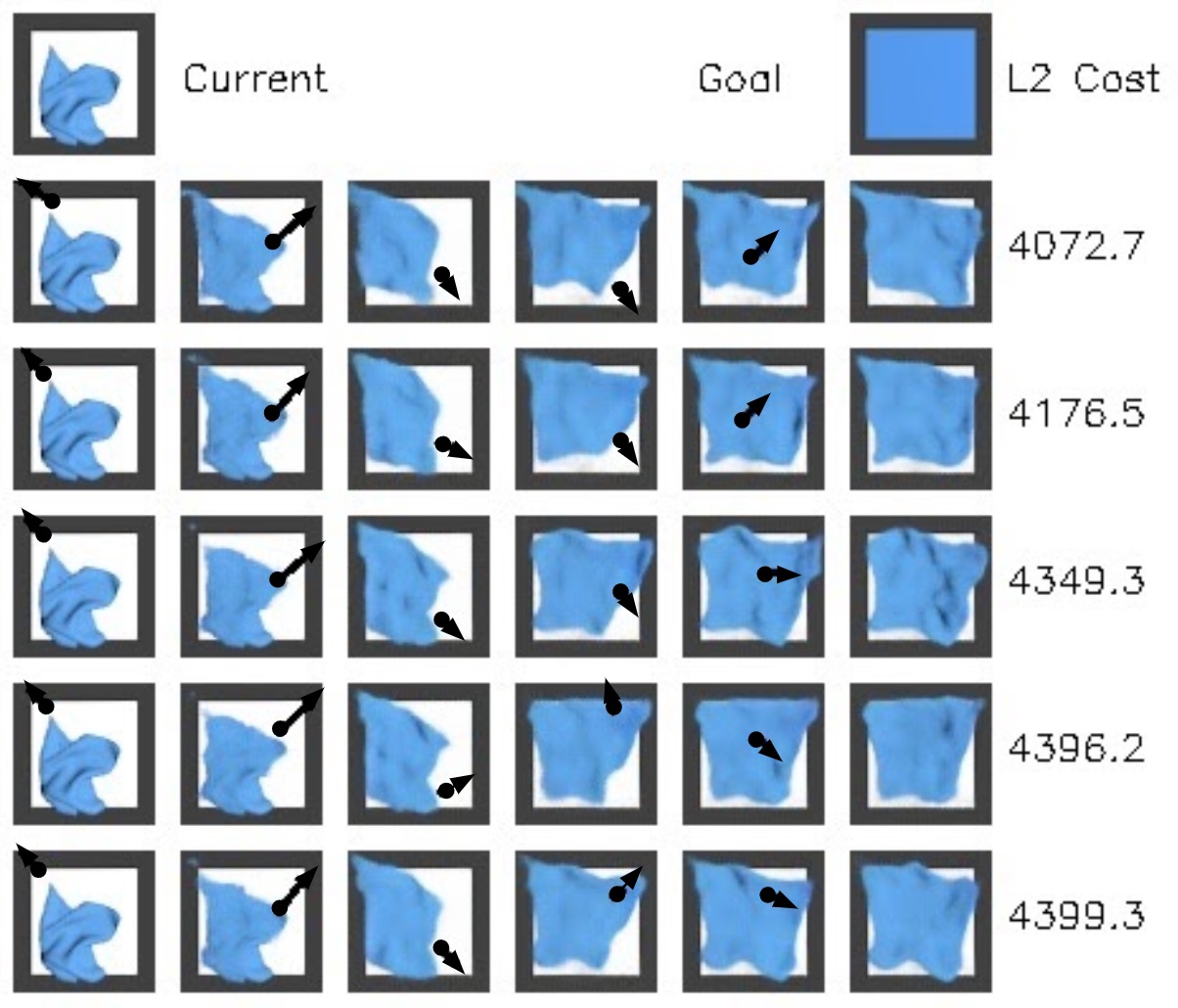}
\caption{
\small
Real plans generated by the system at test time for the smoothing task. We generate action sequences with the Cross Entropy Method (CEM) to approximately minimize the cost function, which evaluates L2 distance between the final image in the predicted trajectory and a provided goal image. Here we show the five CEM trajectories with the lowest cost, where the image sequences in each row are outputs of the video prediction model and the black arrows are the pick-and-pull actions projected onto the images.
}
\vspace*{-10pt}
\label{fig:planning}
\end{figure}

To represent the dynamics of the fabric, we train a deep recurrent convolutional neural network~\cite{Goodfellow-et-al-2016} to predict a sequence of RGBD output frames conditioned on a sequence of RGBD context frames and a sequence of actions. This visuospatial dynamics model is trained on thousands of self-supervised simulated episodes of interaction with the fabric, where an episode consists of a contiguous trajectory of observations and actions. We use Stochastic Variational Video Prediction~\cite{sv2p} as discussed in Section~\ref{ssec:visual}. For planning, we utilize a goal-conditioned planning cost function $c_g(\hat{\bo}_{t+1:t+H})$ with goal $g$, which in the proposed work is a target image $\bo^{(g)}$. The cost is evaluated over the $H$-length sequence of predicted images $\hat{\bo}_{t+1:t+H}$ sampled from the dynamics model conditioned on the current observation $\bo_t$ and some proposed action sequence $\hat{\ba}_{t:t+H-1}$. While there are a variety of cost functions that can be used for visual foresight~\cite{visual_foresight_2018}, we utilize the Euclidean pixel distance between the final predicted RGBD image at timestep $t$ and the goal image $\bo^{(g)}$ across all 4 channels as this works well in practice for the tasks we consider. Precisely, we define the planning cost as follows:
\begin{equation}\label{eq:cost}
    c_g(\hat{\bo}_{t+1:t+i*}) = \|\bo^{(g)} - \hat{\bo}_{t+i*}\|_2
\end{equation}
where
\begin{equation}
    i* = \min \{H, T - t\}
\end{equation}
As in prior work \cite{finn_vf_2017, visual_foresight_2018, robonet}, we utilize the cross entropy method~\cite{cem_1999} to plan action sequences to minimize $c_g(\hat{\bo}_{t+1:t+H})$ over a receding $H$-step horizon at each time $t$. See Figure~\ref{fig:planning} for intuition on the VSF planning phase.


\subsection{Fabric Simulator}\label{ssec:sim}
The fabric and robot simulator we use is built on top of the open-source code from~\citet{seita_ryan}. We briefly review the most relevant aspects of the simulator, with emphasis on the changes from the original.
\subsubsection{Simulator Design} \label{sssec:sim-design}
\vsf requires a large amount of training data to predict full-resolution RGBD images. Since getting real data is cumbersome and imprecise, we use a fabric simulator to generate data quickly and efficiently. We use the fabric simulator from~\citet{seita_ryan}, which was shown to be sufficiently accurate for learning reasonable fabric smoothing policies with imitation learning.

The fabric is represented as a mass-spring system with a $25 \times 25$ grid of point masses~\cite{provot_1996} with springs connecting it to its neighbors. Verlet integration~\cite{verlet_1967} is used to update point mass positions using finite difference approximations and self-collision is implemented by adding a repulsive force between points that are too close~\cite{cloth-cloth-collisions}. Damping is also applied to simulate friction. See Section~\ref{sec:rw} and Appendix~\ref{app:simulators} for further detail on alternate fabric simulators and the simulation used in this work. (The appendix is available on the project website \url{https://sites.google.com/view/fabric-vsf/}).

\subsubsection{Manipulation Details}\label{sssec:manip}

Similar to~\cite{seita_ryan}, actions are represented as four-dimensional vectors as described in Section~\ref{sec:bg} with pick point $(x, y) \in [-1, 1]^2$ and pull vector $(\Delta x, \Delta y) \in [-0.4, 0.4]^2$. We use the open-source software Blender to render (top-down) image observations $\bo_t$ of the fabric. We make several changes for the observations relative to~\cite{seita_ryan}. First, we use four-channel images: three for RGB, and one for depth. Second, we reduce the size of observations to $56\times 56$ from $100\times 100$, to make it more computationally tractable to train visual dynamics models. Finally, to enable transfer of policies trained in simulation to the real-world, we adjust the domain randomization~\cite{domain_randomization} techniques so that color, brightness, and positional hyperparameters are fixed \emph{per episode} to ensure that the video prediction model learns to only focus on predicting changes in the fabric configuration, rather than changes due to domain randomization. See Appendix~\ref{app:results} for more details on the domain randomization parameters.

\subsection{Data Generation}\label{ssec:data-gen}
We collect 7,003 episodes of length 15 each for a total of 105,045 ($\bo_t, \ba_t, \bo_{t+1}$) transitions for training the visuospatial dynamics model. The episode starting states are sampled from three tiers of difficulty with equal probability. These tiers are the same as those in~\cite{seita_ryan}, and largely based on \emph{coverage}, or the amount that a fabric covers its underlying plane:
\begin{itemize}
\item \textbf{Tier 1 (Easy): High Coverage.} $78.3 \pm 6.9$\% initial coverage, all corners visible. Generated by two short random actions. 
\item \textbf{Tier 2 (Medium): Medium Coverage.} $57.6 \pm 6.1$\% initial coverage, one corner occluded. Generated by a vertical drop followed by two actions to hide a corner.
\item \textbf{Tier 3 (Hard): Low Coverage.} $41.1 \pm 3.4$\% initial coverage, 1-2 corners occluded. Generated by executing one action very high in the air and dropping. 
\end{itemize}


All data was generated using the following policy: execute a randomly sampled action, but resample if the grasp point $(x,y)$ is not within the bounding box of the 2D projection of the fabric, and take the additive inverse of $\Delta x$ and/or $\Delta y$ if $(x + \Delta x, y + \Delta y)$ is out of bounds (off the plane by more than 20\% of its width, causing episode termination).

\subsection{VisuoSpatial Dynamics Model}\label{ssec:visual}

Due to the inherent stochasticity in fabric dynamics, we use Stochastic Variational Video Prediction (SV2P) from~\citet{sv2p}, a state-of-the-art method for action-conditioned video prediction. Here, the video prediction model with parameters $\theta$ is designed as a latent variable model, enabling the resulting posterior distribution to capture different modes in the distribution of predicted frames. Precisely,~\cite{sv2p} trains a generative model which predicts a sequence of $H$ output frames conditioned on a context vector of $m$ frames and a sequence of actions starting from the most recent context frame. Since the stochasticity in video prediction is often a consequence of latent events not directly observable in the context frames as noted in \cite{sv2p}, predictions are conditioned on a vector of latent variables $\bz_{t+m:t+m+H-1}$, each sampled from a fixed prior distribution $p(\bz)$. See \citet{sv2p} for more details on the model architecture and training procedure. For this work, we utilize the SV2P implementation provided in~\cite{tensor2tensor}. This gives rise to the following generative model.
\begin{align}
\begin{split}
    &p_\theta(\hat{\bo}_{t+m:t+m+H-1} | \hat{\ba}_{t+m-1:t+m+H-2}, \bo_{t:t+m-1})  = \\ 
    &p(\bz_{t+m}) \prod_{t'= t + m}^{t + m + H - 1} p_\theta(\hat{\bo}_{t'} | \bo_{t:t+m-1}, \hat{\bo}_{t+m:t' - 1}, \bz_{t'}, \hat{\ba}_{t'-1}).
\end{split}
\end{align}
Here $\bo_{t:t+m-1}$ are image observations from time $t$ to $t+m-1$, $\hat{\ba}_{t+m-1:t+m+H-2}$ is a candidate action sequence at timestep $t+m-1$, and $\hat{\bo}_{t+m:t+m+H-1}$ is the sequence of predicted images. Since the generative model is trained in a recurrent fashion, it can be used to sample an $H$-length sequence of predicted images $\hat{\bo}_{t+m:t+m+H-1}$ for any $m > 0, H > 0$ conditioned on a current sequence of image observations $\bo_{t:t+m-1}$ and an $H$-length sequence of proposed actions taken from $\bo_{t+m-1}$, given by $\hat{\ba}_{t+m-1:t+m+H-2}$.

During training, we set the number of context frames $m = 3$ and number of output frames $H = 7$ (the SV2P model learns to predict 7 frames of an episode from the preceding 3 frames). We train on domain-randomized RGBD data, where we randomize fabric color (in a range centered around blue), shading of the plane, image brightness, and camera pose. At test time, we utilize only one context frame $m = 1$ and a planning horizon of $H=5$ output frames. This yields the model $p_\theta(\hat{\bo}_{t+1:t+5} | \hat{\ba}_{t:t+4}, \bo_{t})$.

\subsection{VisuoSpatial Foresight}\label{ssec:vismpc}

To plan over the learned visuospatial dynamics model, we leverage the Cross-Entropy Method (CEM) \cite{cem_1999} as in~\cite{visual_foresight_2018}. CEM repeatedly samples a population of action sequences from a multivariate Gaussian distribution, uses the learned dynamics model to predict output frames for each action sequence, identifies a set of ``elite" sequences with the lowest cost according to the cost function in Section~\ref{subsec:goal-cond}, and refits the Gaussian distribution to these elite sequences. We run 10 iterations of CEM with a population size of 2000 action sequences, 400 elites per iteration, and a planning horizon of 5. See Appendix~\ref{app:visualmpc} for additional hyperparameters. Finally, to mitigate compounding model error we leverage Model-Predictive Control (MPC), which replans at each step and executes only the first action in the lowest cost action sequence found by CEM.
\section{Simulated Experiments}\label{sec:results}


\begin{figure}[t]
\center
\includegraphics[width=0.48\textwidth]{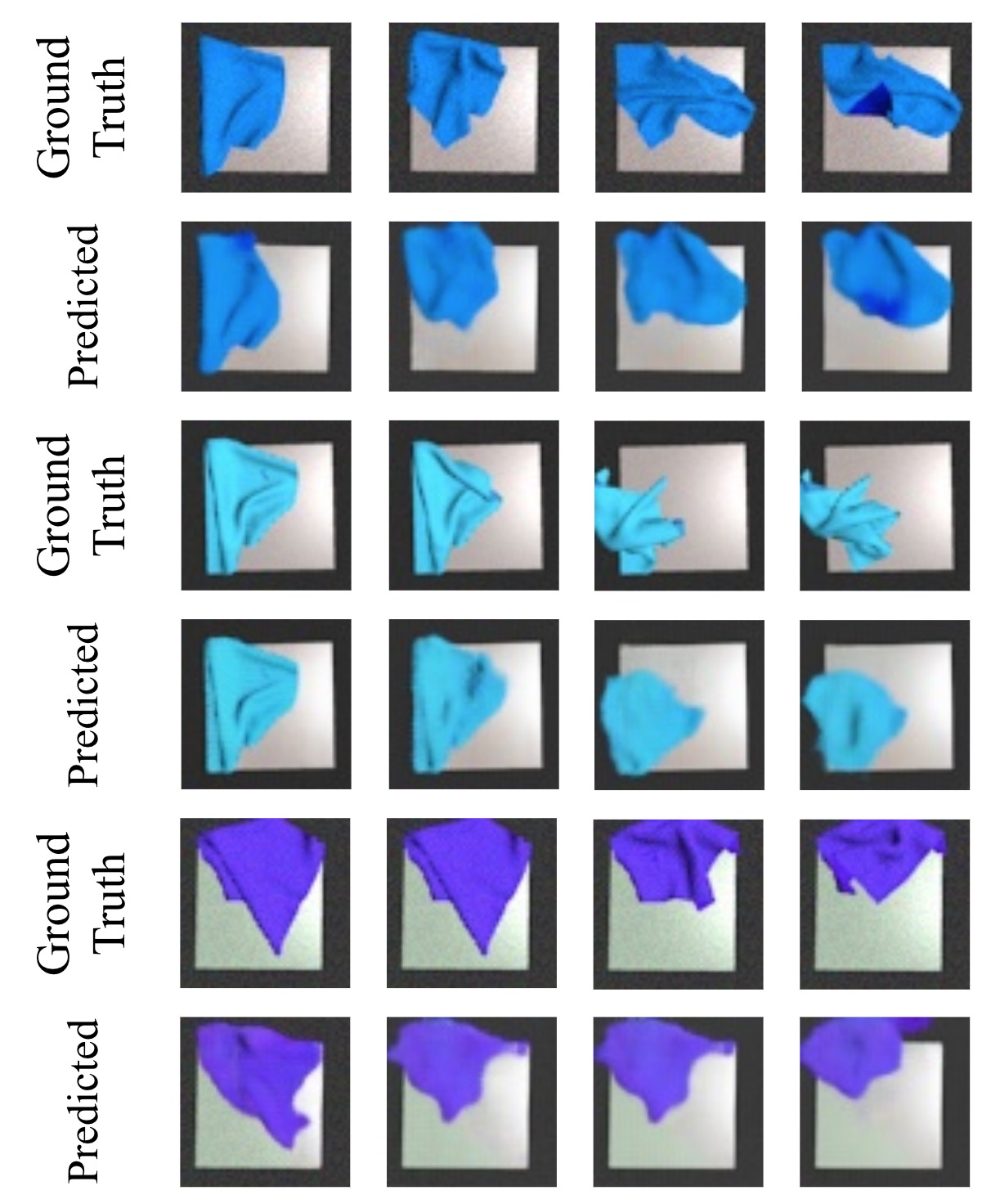}
\caption{
\small
Three pairs of a sequence of four simulated image observations in ground truth compared against a sequence of the corresponding predictions from the visuospatial dynamics model. Here we show the RGB portion of the observations. Each episode is a test example and has randomized camera angle, fabric color, brightness, and shading. Prediction quality varies and generally gets blurrier across time, but is sufficient for planning.
}
\vspace*{-10pt}
\label{fig:dr}
\end{figure}

\subsection{VisuoSpatial Dynamics Prediction Quality}\label{ssec:pred}

One advantage of VSF, and visual foresight more generally, is that we can inspect the model to see if its predictions are accurate, which provides some interpretability. Figure~\ref{fig:dr} shows three examples of predicted image sequences within episodes compared to ground truth images from the actual episodes. All episodes are from test time rollouts of a random policy, meaning that the ground truth images did not appear in the training data for the visuospatial dynamics model (see Sections~\ref{ssec:data-gen} and~\ref{ssec:visual}).

The predictions are reasonably robust to domain randomization, as the model is able to produce fabrics of roughly the appropriate color, and can appropriately align the angle of the light gray background plane. For a more quantitative measure of prediction quality, we calculate the average Structural SIMilarity (SSIM) index~\cite{ssim_2004} over corresponding image pairs in 200 predicted sequences against 200 ground truth sequences to be 0.701. A higher SSIM $\in$ [-1,1] corresponds to higher image similarity.



\subsection{Fabric Smoothing Simulations}\label{ssec:policy}

We first evaluate VSF on the smoothing task: maximizing fabric coverage, defined as the percentage of an underlying plane covered by the fabric. The plane is the same area as the fully smoothed fabric. We evaluate smoothing on three tiers of difficulty as reviewed in Section~\ref{ssec:data-gen}. Each episode lasts up to $T=15$ time steps. Following~\cite{seita_ryan}, episodes can terminate earlier if a threshold of 92\% coverage is triggered, or if any fabric point falls sufficiently outside of the fabric plane.

To see how the general \vsf policy performs against existing smoothing techniques, for each difficulty tier, we execute 200 episodes of \vsf (trained on random, domain-randomized RGBD data) with L2 cost and 200 episodes of each baseline policy in simulation. The L2 cost is taken with respect to a smooth goal image (see Figure~\ref{fig:rollout_smooth}). Note that VSF does \textit{not} explicitly optimize for coverage, and only optimizes the cost function from Equation~\ref{eq:cost}, which measures L2 pixel distance to a target image. We compare with the following 5 baselines:

\begin{figure*}[t]
\center
\includegraphics[width=0.95\textwidth]{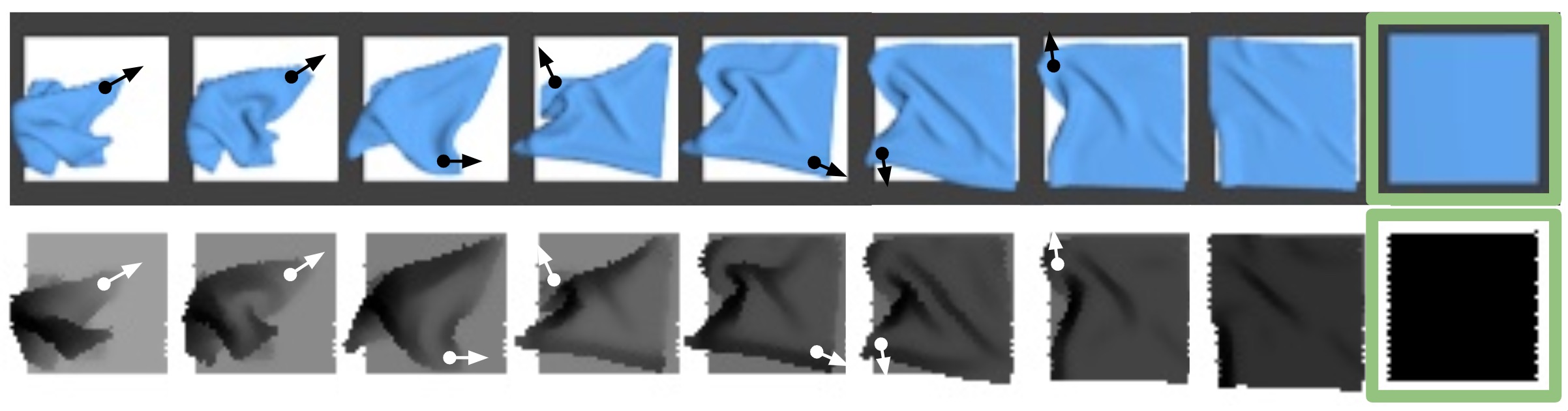}
\caption{
\small
A simulated episode executed by the \vsf policy on a Tier 3 starting state, given a smooth goal image (shown in the far right). The first row shows RGB images and the second shows the corresponding depth maps. In this example, the policy is able to successfully cross the coverage threshold of 92\% after executing 7 actions. Actions are visualized with the overlaid arrows.
}
\vspace*{-10pt}
\label{fig:rollout_smooth}
\end{figure*}

\paragraph{Random} Random pick point and pull direction.
\paragraph{Highest} Using ground truth state information, pick the point mass with the maximum $z$-coordinate and set pull direction to point to where the point mass would be if the fabric were perfectly smooth. This is straightforward to implement with depth sensing and was shown to work reasonably well for smoothing in~\cite{seita-bedmaking}.
\paragraph{Wrinkle} As in~\citet{cloth_icra_2015}, find the largest wrinkle and then pull perpendicular to it at the edge of the fabric to smooth it out. We use the ground truth state information in the implementation of this algorithm (as done in~\cite{seita_ryan}) rather than image observations.
\paragraph{Imitation Learning (IL)} As in~\citet{seita_ryan}, train an imitation learning agent using DAgger~\cite{ross2011reduction} with a simulated ``corner-pulling" demonstrator that picks and pulls at the fabric corner furthest from its target. DAgger can be considered as an oracle with ``privileged'' information as in~\citet{cheating_2019} because during training, it queries a demonstrator which uses ground truth state information. For fair comparison, we run DAgger so that it consumes roughly the same number of data points (we used 110,000) as \vsf during training, and we give the policy access to four-channel RGBD images. We emphasize that this is a distinct dataset from the one used for VSF, which uses no demonstrations.
\paragraph{Model-Free RL} We run DDPG~\cite{ddpg2016} and extend it to use demonstrations and a pre-training phase as suggested in~\citet{ddpgfd}. We also use the Q-filter from~\citet{overcoming_exploration}. We train with a similar number of data points as in IL and VSF for a reasonable comparison. We design a reward function for the smoothing task that, at each time step, provides reward equal to the change in coverage between two consecutive states. Inspired by~\citet{openai-dactyl}, we provide a $+5$ bonus for triggering a coverage success, and $-5$ penalty for pulling the fabric out of bounds.

Further details about the implementation and training of IL, DDPG, and VSF are in Appendix~\ref{app:implementation}.

\begin{table}[t]
\small
\caption{
\small
Simulated smoothing experimental results for the six policies in Section~\ref{ssec:policy}. We report final coverage and number of actions per episode, averaged over 200 simulated episodes per tier. \vsf (VSF) performs well even for difficult starting states. It attains similar final coverage as the \il (IL) agent from~\cite{seita_ryan} and outperforms the other baselines. The VSF and IL agents were trained on equal amounts of domain-randomized RGBD data, but the IL agent has a demonstrator for every training state, whereas VSF is trained with data collected from a random policy.
}
\centering
\begin{tabular}{l l | r r}
\textbf{Tier} & \textbf{Method} & \textbf{Coverage} & \textbf{Actions} \\ \hline 
1 & Random  &  25.0 $\pm$ 14.6 & 2.4 $\pm$ 2.2  \\
1 & Highest &  66.2 $\pm$ 25.1 & 8.2 $\pm$ 3.2  \\
1 & Wrinkle &  91.3 $\pm$ \:\:7.1  & 5.4 $\pm$ 3.7  \\
1 & DDPG and Demos & 87.1 $\pm$ 10.7 & 8.7 $\pm$ 6.1 \\
1 & \il & 94.3 $\pm$ \:\:2.3  & 3.3 $\pm$ 3.1 \\
1 & \vsf & 92.5 $\pm$ \:\:2.5 & 8.3 $\pm$ 4.7 \\ \hline
2 & Random  &  22.3 $\pm$ 12.7  & 3.0 $\pm$ 2.5  \\ 
2 & Highest &  57.3 $\pm$ 13.0  & 10.0 $\pm$ 0.3  \\
2 & Wrinkle &  87.0 $\pm$ 10.8  & 7.6 $\pm$ 2.8  \\
2 & DDPG and Demos & 82.0 $\pm$ 14.7 & 9.5 $\pm$ 5.8 \\
2 & \il &  92.8 $\pm$ \:\:7.0  &  5.7 $\pm$ 4.0 \\
2 & \vsf & 90.3 $\pm$ \:\:3.8 & 12.1 $\pm$ 3.4 \\ \hline
3 & Random  &  20.6 $\pm$ 12.3  & 3.8 $\pm$ 2.8  \\ 
3 & Highest &  36.3 $\pm$ 16.3  & 7.9 $\pm$ 3.2  \\ 
3 & Wrinkle &  73.6 $\pm$ 19.0  & 8.9 $\pm$ 2.0  \\ 
3 & DDPG and Demos & 67.9 $\pm$ 15.6 & 12.9 $\pm$ 3.9 \\
3 & \il &  88.6 $\pm$ 11.5 & 10.1 $\pm$ 3.9  \\
3 & \vsf & 89.3 $\pm$ \:\:5.9 & 13.1 $\pm$ 2.9 \\
\end{tabular}
\vspace*{-5pt}
\label{tab:analytic}
\end{table}

Table~\ref{tab:analytic} indicates that VSF significantly outperforms the analytic and model-free reinforcement learning baselines. It has similar performance to the IL agent, a ``smoothing specialist'' that rivals the performance of the corner pulling demonstrator used in training (see Appendix~\ref{app:implementation}). See Figure~\ref{fig:rollout_smooth} for an example Tier 3 VSF episode. Furthermore, we find coverage values to be statistically significant compared to all baselines other than IL and not much impacted by the presence of domain randomization. Results from the Mann-Whitney U test~\cite{mannwhitney} and a domain randomization ablation study are reported in Appendix~\ref{app:results}. VSF, however, requires more actions than DAgger, especially on Tier 1, with 8.3 actions per episode compared to 3.3 per episode. We hypothesize that this leads to VSF being more conservative and reliable, which is supported by VSF having lower variance in performance on Tier 2 and Tier 3 settings (Table~\ref{tab:analytic}).
\begin{table}[t]
\small
\caption{
\small
Single folding results in simulation. \vsf is run with the goal image in Figure~\ref{fig:rollout_folding} for 20 episodes when L2 is taken on the depth, RGB, and RGBD channels. The results suggest that adding depth allows us to significantly outperform RGB-only Visual Foresight on this task.
}
\centering
\begin{tabular}{l | r r r}
\textbf{Cost Function} & \textbf{Successes} & \textbf{Failures} & \textbf{\% Success} \\ \hline 
L2 Depth  & 0 & 20 & 0\%  \\
L2 RGB  & 10 & 10 & 50\%  \\
\textbf{L2 RGBD}  & \textbf{18} & \textbf{2} & \textbf{90\%}  \\
\end{tabular}
\vspace*{-5pt}
\label{tab:folding_trials}
\end{table}

\begin{figure}[t]
\center
\includegraphics[width=0.35\textwidth]{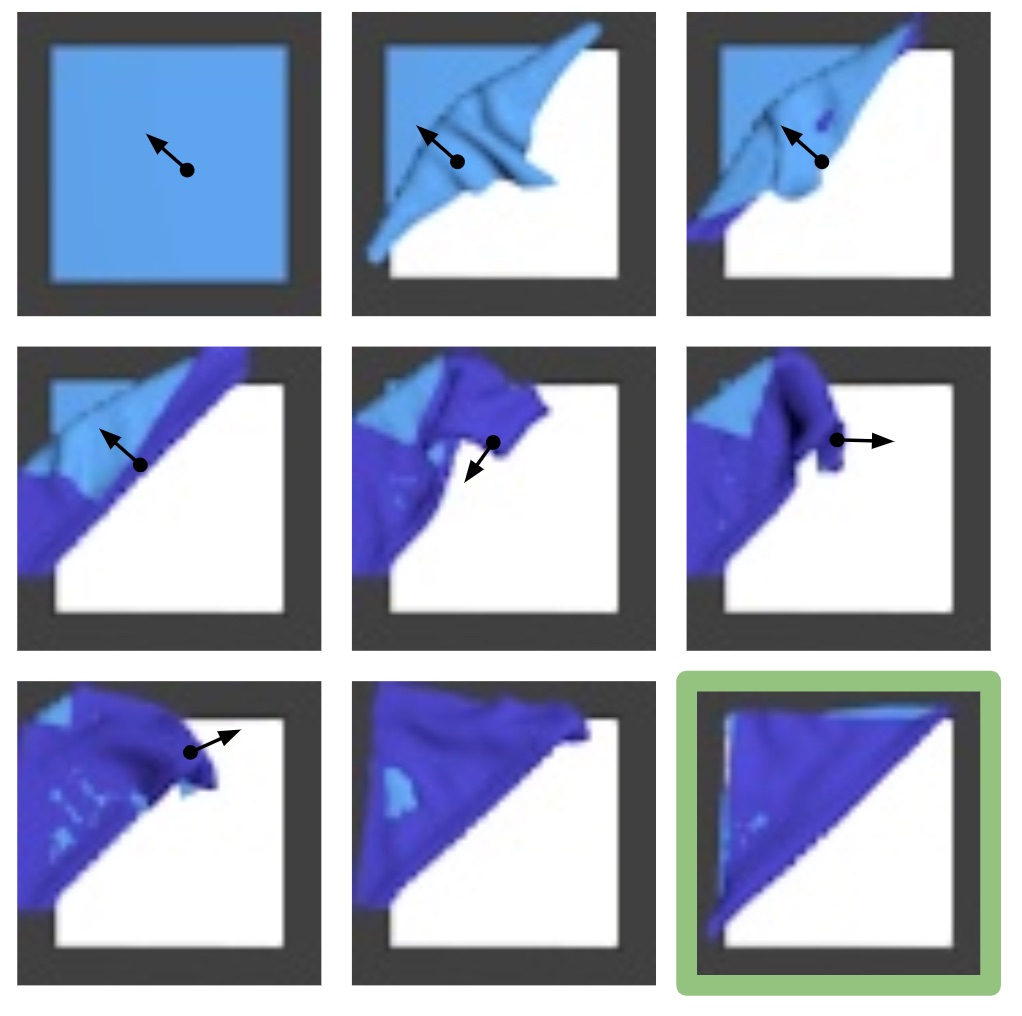}
\caption{
\small
Simulated RGB observations of a successful folding episode. The RGB portion of the goal image is displayed in the bottom right corner. The rest is an 8-step episode (left-to-right, top-to-bottom) from smooth to approximately folded. There are several areas of the fabric simulator which have overlapping layers due to the difficulty of accurately modeling fabric-fabric collisions in simulation, which explain the light blue patches in the figure.
}
\vspace*{-10pt}
\label{fig:rollout_folding}
\end{figure}

\begin{figure}[t]
\center
\includegraphics[width=0.35\textwidth]{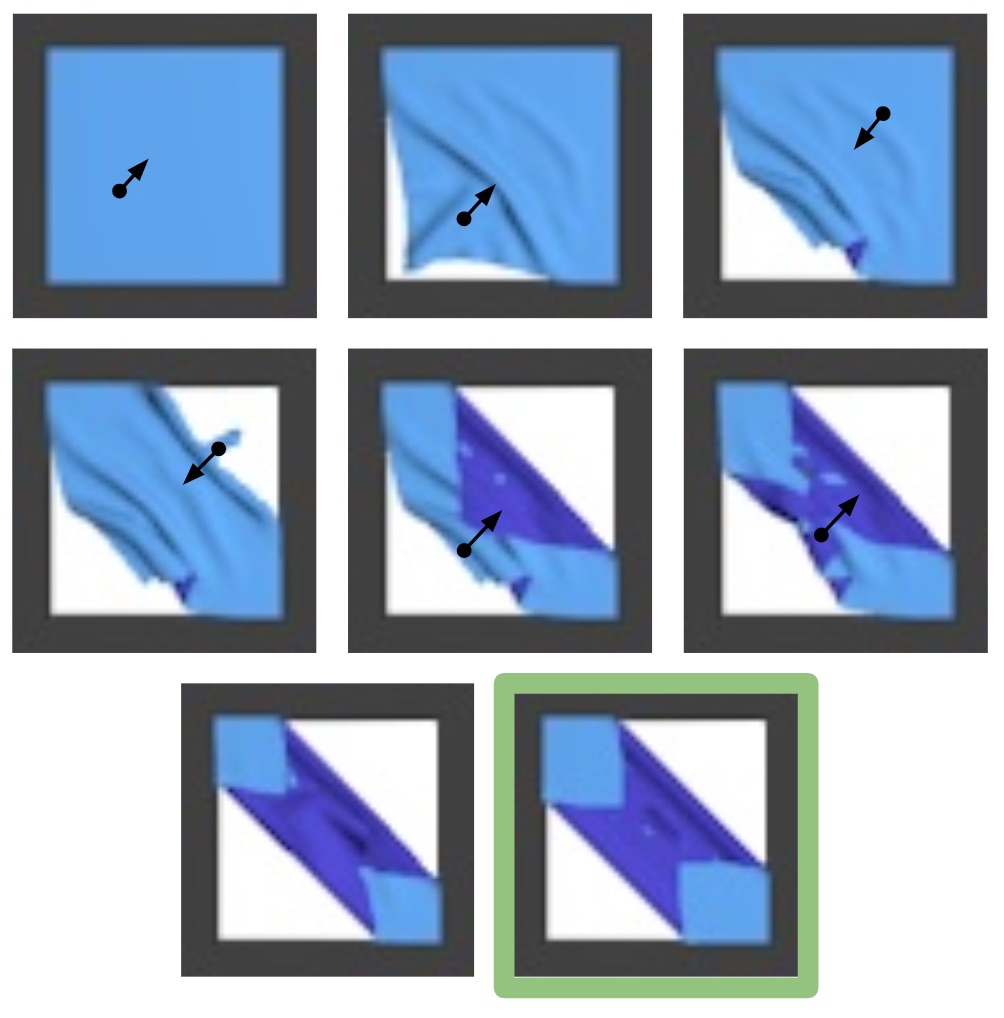}
\caption{
\small
An example of a successful double folding episode. The RGB portion of the goal image is displayed in the bottom row. The rest is an 7-step trajectory (left-to-right, top-to-bottom) from smooth to approximately matching the goal image. The two folds are stacked in the correct order.
}
\vspace*{-10pt}
\label{fig:two_folds}
\end{figure}

\subsection{Fabric Folding Simulations}\label{ssec:folding}

Here we demonstrate that VSF is not only able to smooth fabric, but also directly generalizes to folding tasks. We use the same video prediction model, trained only with random interaction data, and keep planning parameters the same besides the initial CEM variance (see Appendix~\ref{app:visualmpc}). We change the goal image to the triangular, folded shape shown in Figure~\ref{fig:rollout_folding} and change the initial state to a smooth state (which can be interpreted as the result of a smoothing policy from Section~\ref{ssec:policy}). The two sides of the fabric are shaded differently, with the darker shade on the bottom layer. Due to the action space bounds (Section~\ref{sssec:manip}), getting to this goal state directly is not possible in under three actions and requires a precise sequence of pick and pull actions.

\begin{figure}[t]
\center
\includegraphics[width=0.4\textwidth]{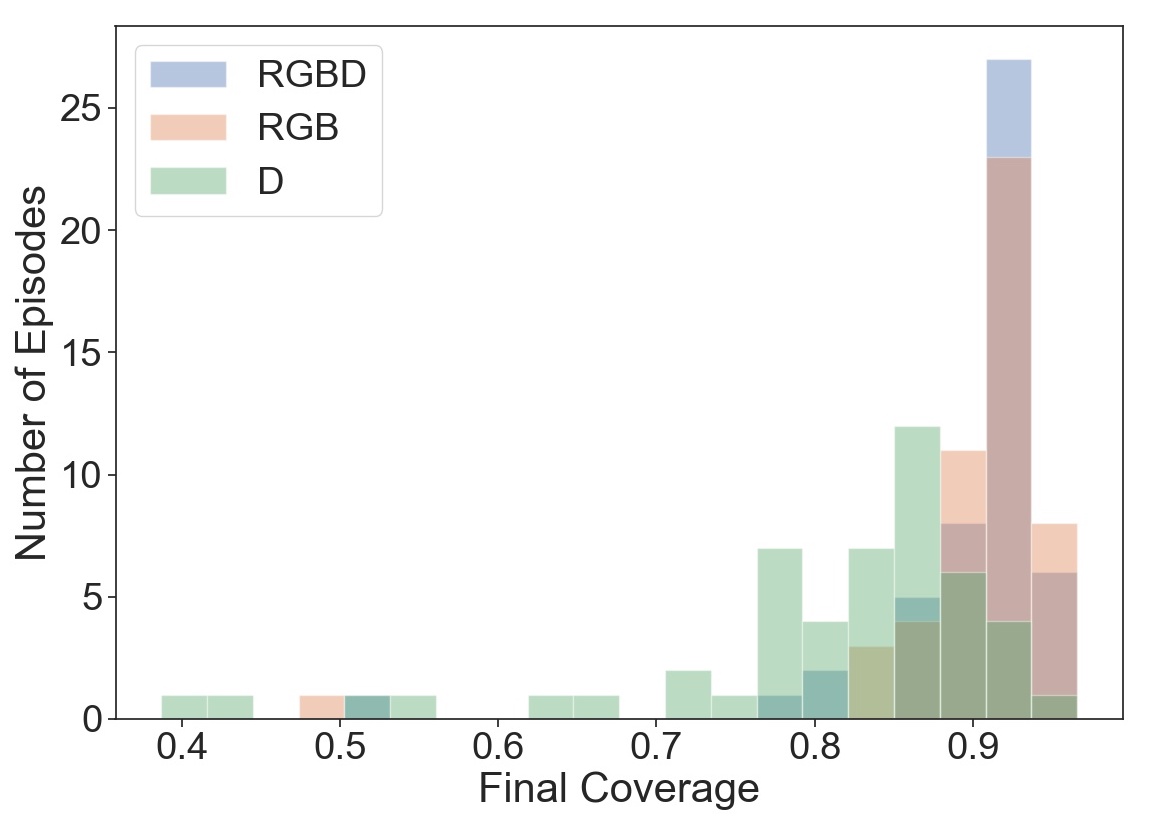}
\caption{
\small
Final coverage values on 50 simulated smoothing episodes from Tier 3 starting states. We fix the random seed so that each input modality (RGB, D and RGBD) begins with the same starting states.
}
\vspace*{-10pt}
\label{fig:hist}
\end{figure}

We visually inspect the final states in each episode, and classify them as successes or failures. For RGBD images, this decision boundary empirically corresponds to an L2 threshold of roughly 2500; see Figure~\ref{fig:rollout_folding} for a typical success case. In Table~\ref{tab:folding_trials} we compare performance of L2 cost taken over RGB, depth, and RGBD channels. RGBD significantly outperforms the other modes, which correspond to Visual Foresight and ``Spatial Foresight'' respectively, suggesting the usefulness of augmenting Visual Foresight with depth maps. For a more complete analysis of the impact of input modality, we also provide a histogram of coverage values obtained on the simulated \textit{smoothing} task for RGB, D, and RGBD in Figure~\ref{fig:hist}. Here RGBD also performs the best but only slightly outperforms RGB, which is perhaps unsurprising due to the relatively low depth variation in the smoothing task.

We then attempt to reach a more complex goal image with two overlapping folds for a total of three layers at the center of the fabric, again with the same VSF policy. Here 8 out of 20 trials succeed, and 7 of the 8 successes arrange the two folds in the correct ordering. Folding two opposite corners in the right order may be difficult to do solely with RGB data, since the bottom layer of the fabric has a uniform color. The depth maps, in contrast, provide information about one fold lying on top of another. See Figure~\ref{fig:two_folds} for an example of a successful rollout. Failure cases are often characterized by the robot picking at a point that just misses a fabric corner, but where the visual dynamics model predicts that the resulting pull will behave as if the robot had picked at the fabric corner.

\section{Physical Experiments}\label{sec:physical-results}

We next deploy the domain randomized policies on a physical da Vinci surgical robot~\cite{dvrk2014}. We use the same experimental setup as in~\citet{seita_ryan}, including the calibration procedure to map pick points $(x,y)$ into positions and orientations with respect to the robot's base frame. The sequential tasks we consider are challenging due to the robot's imprecision~\cite{seita_icra_2018}. We use a Zivid One Plus camera mounted 0.9 meters above the workspace to capture RGBD images.

\begin{table}[t]
\small
\caption{
\small
Physical smoothing robot experiment results for \il (IL), i.e. DAgger, and \vsf (VSF). For both methods, we choose the policy snapshot with highest performance in simulation, and each are applied on all tiers (T1, T2, T3). We show results across 10 episodes of IL per tier and 5 episodes of VSF per tier, and show average starting and final coverage, maximum coverage at any point per episode, and the number of actions. Results suggest that VSF attains final coverage comparable to or exceeding that of IL despite not being trained on demonstration data, though VSF requires more actions per episode.
}
\centering
\begin{tabular}{l | l l l r }
Tier & (1) Start & (2) Final & (3) Max & (4) Actions \\ \hline 
1 IL & 74.2 $\pm$ 5 & 92.1 $\pm$ \:\:6 & 92.9 $\pm$ \:\:3 & 4.0 $\pm$ 3 \\
1 VSF & 78.3 $\pm$ 6 & \textbf{93.4 $\pm$ \:\:2} & 93.4 $\pm$ \:\:2 & 8.2 $\pm$ 4 \\ \hline
2 IL & 58.2 $\pm$ 3 & 84.2 $\pm$ 18 & 86.8 $\pm$ 15 & 9.8 $\pm$ 5 \\
2 VSF & 59.5 $\pm$ 3 & \textbf{87.1 $\pm$ \:\:9} & 90.0 $\pm$ \:\:5 & 12.8 $\pm$ 3 \\ \hline
3 IL & 43.3 $\pm$ 4 & 75.2 $\pm$ 18 & 79.1 $\pm$ 14 & 12.5 $\pm$ 4 \\
3 VSF & 41.4 $\pm$ 3 & \textbf{75.6 $\pm$ 15} & 76.9 $\pm$ 15 & 15.0 $\pm$ 0 \\
\end{tabular}
\vspace*{-5pt}
\label{tab:surgical}
\end{table}


\subsection{Experiment Protocol}\label{ssec:protocol}

We evaluate the best \il and the best \vsf policies as measured in simulation and reported in Table~\ref{tab:analytic}. We do not test with the model-free DDPG policy baseline, as it performed significantly worse than the other two methods. For IL, this is the final model trained with 110,000 actions based on a corner-pulling demonstrator with access to state information. This uses slightly more than the 105,045 actions used for training the VSF model. We reiterate that VSF uses a video prediction model that is trained on \emph{entirely random data}, requiring no labeled data in contrast with IL, which uses DAgger and requires a demonstrator with ``privileged'' fabric state information during training. The demonstrator uses state information to determine the fabric corner furthest from its target on the underlying plane, and pulls the fabric at that fabric corner to the target.

\begin{figure*}[t]
\center
\includegraphics[width=0.90\textwidth]{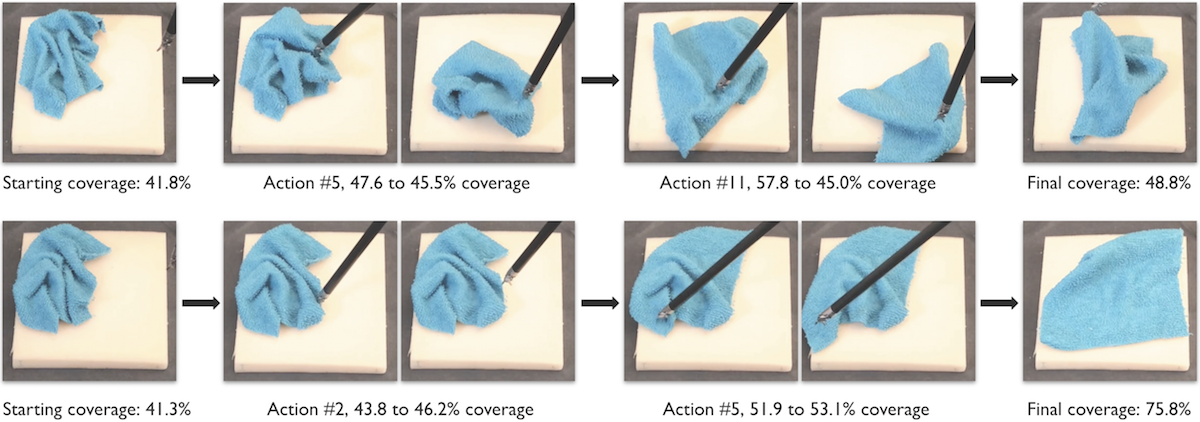}
\caption{
\small
A qualitative comparison of physical da Vinci episodes with an \il policy (top row) and a \vsf policy (bottom row). The rows show screen captures taken from the third-person video view for recording episodes; these are not the input to \vsf. To facilitate comparisons among IL and VSF, we manually make the starting fabric state as similar as possible. Over the course of several actions, the IL policy sometimes takes actions that are highly counter-productive, such as the 5th and 11th actions above. Both pick points are reasonably chosen, but the large deltas cause the lower right fabric corner to get hidden. In contrast, VSF, takes shorter pulls on average, with representative examples shown above for the 2nd and 5th actions. At the end, the IL policy gets just 48.8\% coverage (far below its usual performance), whereas VSF gets 75.8\%. For further quantitative results, see Table~\ref{tab:surgical}.
}
\vspace*{-10pt}
\label{fig:il_vs_vf_example}
\end{figure*}

To match the simulation setup, we limit each episode to a maximum of 15 actions. For both methods on the \emph{smoothing} task, we perform episodes in which the fabric is initialized in highly rumpled states which mirror those from the simulated tiers. We run ten episodes per tier for IL and five episodes per tier for VSF, for 45 episodes in all. In addition, within each tier, we attempt to make starting fabric states reasonably comparable among IL and VSF episodes (e.g., see Figure~\ref{fig:il_vs_vf_example}).

\subsection{Physical Fabric Smoothing}\label{ssec:physical-smoothing}

We present quantitative results in Table~\ref{tab:surgical} that suggest that VSF gets final coverage results comparable to that of IL, despite not being trained on any corner-pulling demonstration data. However, it sometimes requires more actions to complete an episode and takes significantly more computation time, as the Cross Entropy Method requires thousands of forward passes through a large network while IL requires only a single pass. 
The actions for VSF usually have smaller deltas, which helps to take more precise actions. This is likely due in part to the initialization of the mean and variance used to fit the conditional Gaussians for CEM. Figure~\ref{fig:action_deltas} in Appendix~\ref{app:act-mag} shows histograms of the action delta magnitudes.
The higher magnitude actions of the IL policy may cause it to be more susceptible to highly counter-productive actions. The fabric manipulation tasks we consider require high precision, and a small error in the pick point region coupled with a long pull may cover a corner or substantially decrease coverage.

As an example, Figure~\ref{fig:il_vs_vf_example} shows a time lapse of a subset of actions for one episode from IL and VSF. Both begin with a fabric of roughly the same shape to facilitate comparisons. On the fifth action, the IL policy has a pick point that is slightly north of the ideal spot. The pull direction to the lower right fabric plane corner is reasonable, but due to the length of the pull, combined with a slightly suboptimal pick point, the lower right fabric corner gets covered. This makes it harder for a policy trained from a corner-pulling demonstrator to get high coverage, as the fabric corner is hidden. In contrast, the VSF policy takes actions of shorter magnitudes and does not fall into this trap. The downside of VSF, however, is that it may ``waste'' too many actions with short magnitudes, whereas IL can quickly get high coverage conditioned on accurate pick points. In future work, we will develop ways to tune VSF so that it takes longer actions as needed.

\subsection{Physical Fabric Folding}\label{ssec:qualitative}


\begin{figure}[t]
\center
\includegraphics[width=0.45\textwidth]{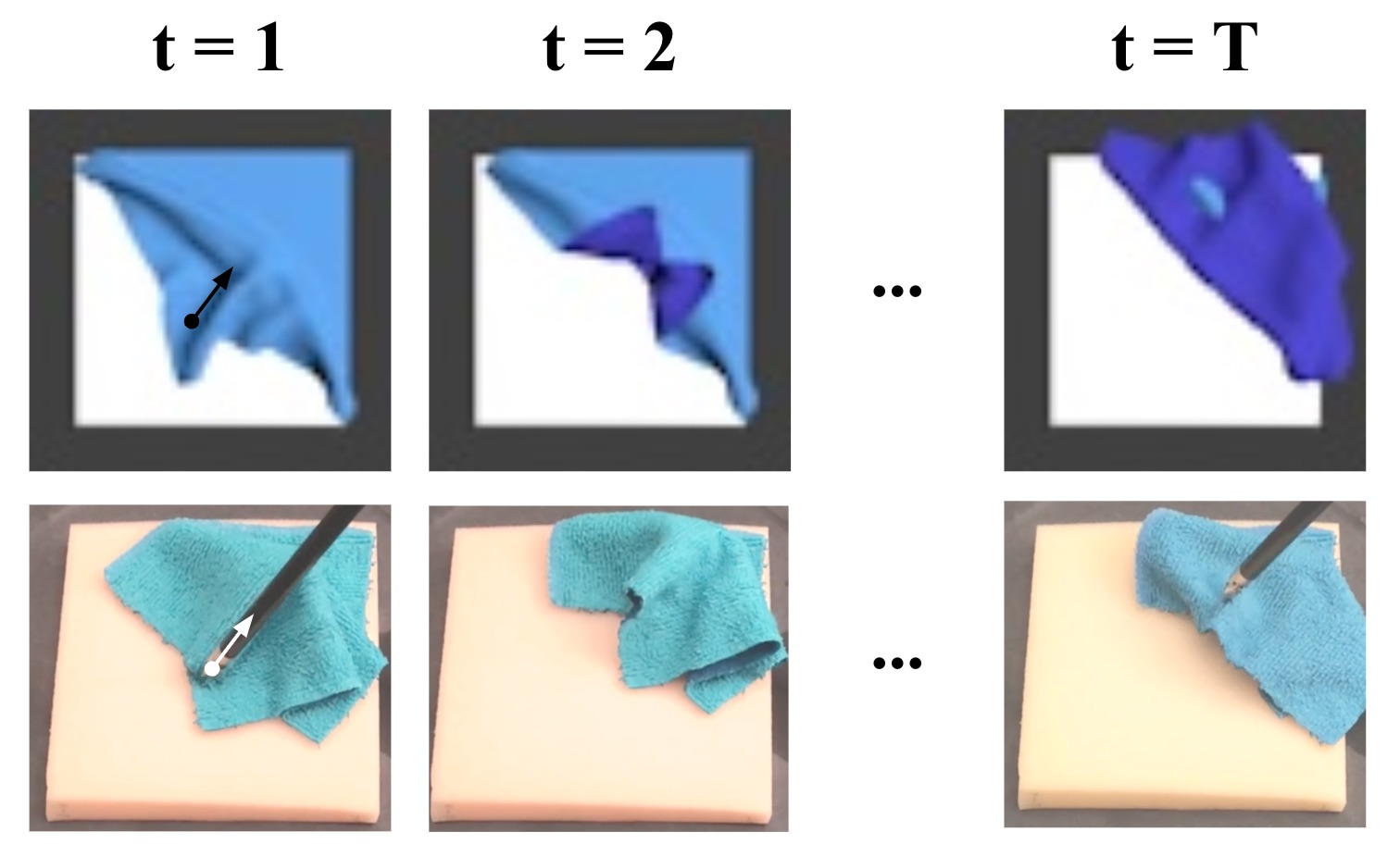}
\caption{
\small
Physical folding policy executed in the simulator and on the surgical robot, with actions determined from \emph{real} image input only. Despite this, the actions are able to fold in simulation. The difference in dynamics is apparent from $t=1$ to $t=2$, where the simulated fabric's bottom left corner is overturned from the action, but the corresponding corner on the real fabric is not.
}
\vspace*{-10pt}
\label{fig:simtoreal}
\end{figure}

We next evaluate the \emph{same} VSF policy, but on a fabric \emph{folding} task, starting from a smooth state. We conduct four goal-conditioned physical folding episodes, with one episode for each of the four possible diagonal folds. We run each episode for 15 time steps. Qualitatively, the robot tends to move the fabric in the right direction based on the target corner, but does not get the fabric in a clean two-layer fold. To evaluate the quality of the policy, we take the actions generated for a \textit{real episode} on the physical system and run them open-loop in the fabric simulator. See Figure~\ref{fig:simtoreal} for a comparison. Since the \textit{same actions} are able to fold reasonably well in simulation, we conclude that the difference is due to a dynamics mismatch between simulated and real environments, compounded with pick point inaccuracies common in cable-driven robots such as the dVRK~\cite{seita_icra_2018}. In future work, we will improve the simulator's physics to more closely match those of the dVRK, and we will also consider augmenting the video prediction model with a small amount of real-world data.

\section{Conclusion and Future Work}\label{sec:conclusions}

We present VSF, which leverages a combination of RGB and depth information to learn goal conditioned fabric manipulation policies for a variety of long horizon tasks. We train a video prediction model on purely random interaction data with fabric in simulation, and demonstrate that planning over this model with MPC results in a policy with promising generalization across goal-directed fabric smoothing and folding tasks. We then transfer this policy to a real robot system with domain randomization. In future work, we will explore learned cost functions, test different fabric shapes, and investigate learning bilateral manipulation or human-in-the-loop policies.


\section*{Acknowledgments}
{\small
This research was performed at the AUTOLAB at UC Berkeley in affiliation with Honda Research Institute USA, the Berkeley AI Research (BAIR) Lab, Berkeley Deep Drive (BDD), the Real-Time Intelligent Secure Execution (RISE) Lab, and the CITRIS ``People and Robots'' (CPAR) Initiative, and by the Scalable Collaborative Human-Robot Learning (SCHooL) Project, NSF National Robotics Initiative Award 1734633. The authors were supported in part by Siemens, Google, Amazon Robotics, Toyota Research Institute, Autodesk, ABB, Samsung, Knapp, Loccioni, Intel, Comcast, Cisco, Hewlett-Packard, PhotoNeo, NVidia, and Intuitive Surgical. Daniel Seita is supported by a National Physical Science Consortium Fellowship and Ashwin Balakrishna is supported by an NSF GRFP.
}

\small
\bibliographystyle{plainnat}
\bibliography{example}

\begin{thebibliography}{66}
\providecommand{\natexlab}[1]{#1}
\providecommand{\url}[1]{\texttt{#1}}
\expandafter\ifx\csname urlstyle\endcsname\relax
  \providecommand{\doi}[1]{doi: #1}\else
  \providecommand{\doi}{doi: \begingroup \urlstyle{rm}\Url}\fi

\bibitem[Babaeizadeh et~al.(2018)Babaeizadeh, Finn, Erhan, Campbell, and
  Levine]{sv2p}
Mohammad Babaeizadeh, Chelsea Finn, Dumitri Erhan, Roy~H. Campbell, and Sergey
  Levine.
\newblock {Stochastic Variational Video Prediction}.
\newblock In \emph{International Conference on Learning Representations
  (ICLR)}, 2018.

\bibitem[Balaguer and Carpin(2011)]{balaguer2011combining}
Benjamin Balaguer and Stefano Carpin.
\newblock {Combining Imitation and Reinforcement Learning to Fold Deformable
  Planar Objects}.
\newblock In \emph{IEEE/RSJ International Conference on Intelligent Robots and
  Systems (IROS)}, 2011.

\bibitem[Balakrishna et~al.(2019)Balakrishna, Thananjeyan, Lee, Zahed, Li,
  Gonzalez, and Goldberg]{converging-supervisor}
Ashwin Balakrishna, Brijen Thananjeyan, Jonathan Lee, Arsh Zahed, Felix Li,
  Joseph~E. Gonzalez, and Ken Goldberg.
\newblock {On-Policy Robot Imitation Learning from a Converging Supervisor}.
\newblock In \emph{Conference on Robot Learning (CoRL)}, 2019.

\bibitem[Baraff and Witkin(1998)]{cloth-cloth-collisions}
David Baraff and Andrew Witkin.
\newblock {Large Steps in Cloth Simulation}.
\newblock In \emph{ACM SIGGRAPH}, 1998.

\bibitem[Berkenkamp et~al.(2016)Berkenkamp, Schoellig, and
  Krause]{berkenkamp2016safe}
Felix Berkenkamp, Angela~P Schoellig, and Andreas Krause.
\newblock {Safe Controller Optimization for Quadrotors with Gaussian
  Processes}.
\newblock In \emph{IEEE International Conference on Robotics and Automation
  (ICRA)}, 2016.

\bibitem[Borras et~al.(2019)Borras, Alenya, and
  Torras]{grasp_centered_survey_2019}
Julia Borras, Guillem Alenya, and Carme Torras.
\newblock {A Grasping-centered Analysis for Cloth Manipulation}.
\newblock \emph{arXiv:1906.08202}, 2019.

\bibitem[Chen et~al.(2019)Chen, Zhou, Koltun, and Krahenbuhl]{cheating_2019}
Dian Chen, Brady Zhou, Vladlen Koltun, and Philipp Krahenbuhl.
\newblock {Learning by Cheating}.
\newblock In \emph{Conference on Robot Learning (CoRL)}, 2019.

\bibitem[Chiuso and Pillonetto(2019)]{system-id}
A.~Chiuso and G.~Pillonetto.
\newblock {System Identification: A Machine Learning Perspective}.
\newblock \emph{Annual Review of Control, Robotics, and Autonomous Systems},
  2019.

\bibitem[Chua et~al.(2018)Chua, Calandra, McAllister, and
  Levine]{handful-of-trials}
Kurtland Chua, Roberto Calandra, Rowan McAllister, and Sergey Levine.
\newblock {Deep Reinforcement Learning in a Handful of Trials Using
  Probabilistic Dynamics Models}.
\newblock In \emph{Neural Information Processing Systems (NeurIPS)}, 2018.

\bibitem[Community(2018)]{blender}
Blender~Online Community.
\newblock \emph{Blender - a 3D modelling and rendering package}.
\newblock Blender Foundation, Stichting Blender Foundation, Amsterdam, 2018.
\newblock URL \url{http://www.blender.org}.

\bibitem[Coumans and Bai(2016--2019)]{coumans2019}
Erwin Coumans and Yunfei Bai.
\newblock Pybullet, a python module for physics simulation for games, robotics
  and machine learning.
\newblock \url{http://pybullet.org}, 2016--2019.

\bibitem[Dasari et~al.(2019)Dasari, Ebert, Tian, Nair, Bucher, Schmeckpeper,
  Singh, Levine, and Finn]{robonet}
Sudeep Dasari, Frederik Ebert, Stephen Tian, Suraj Nair, Bernadette Bucher,
  Karl Schmeckpeper, Siddharth Singh, Sergey Levine, and Chelsea Finn.
\newblock {RoboNet: Large-Scale Multi-Robot Learning}.
\newblock In \emph{Conference on Robot Learning (CoRL)}, 2019.

\bibitem[Dhariwal et~al.(2017)Dhariwal, Hesse, Klimov, Nichol, Plappert,
  Radford, Schulman, Sidor, Wu, and Zhokhov]{baselines}
Prafulla Dhariwal, Christopher Hesse, Oleg Klimov, Alex Nichol, Matthias
  Plappert, Alec Radford, John Schulman, Szymon Sidor, Yuhuai Wu, and Peter
  Zhokhov.
\newblock {OpenAI Baselines}.
\newblock \url{https://github.com/openai/baselines}, 2017.

\bibitem[Doumanoglou et~al.(2014)Doumanoglou, Kargakos, Kim, and
  Malassiotis]{unfolding_rf_2014}
Andreas Doumanoglou, Andreas Kargakos, Tae-Kyun Kim, and Sotiris Malassiotis.
\newblock {Autonomous Active Recognition and Unfolding of Clothes Using Random
  Decision Forests and Probabilistic Planning}.
\newblock In \emph{IEEE International Conference on Robotics and Automation
  (ICRA)}, 2014.

\bibitem[Ebert et~al.(2018)Ebert, Finn, Dasari, Xie, Lee, and
  Levine]{visual_foresight_2018}
Frederik Ebert, Chelsea Finn, Sudeep Dasari, Annie Xie, Alex Lee, and Sergey
  Levine.
\newblock {Visual Foresight: Model-Based Deep Reinforcement Learning for
  Vision-Based Robotic Control}.
\newblock \emph{arXiv:1812.00568}, 2018.

\bibitem[Erickson et~al.(2018)Erickson, Clever, Turk, Liu, and
  Kemp]{deep_dressing_2018}
Zackory Erickson, Henry~M. Clever, Greg Turk, C.~Karen Liu, and Charles~C.
  Kemp.
\newblock {Deep Haptic Model Predictive Control for Robot-Assisted Dressing}.
\newblock In \emph{IEEE International Conference on Robotics and Automation
  (ICRA)}, 2018.

\bibitem[Finn and Levine(2017)]{finn_vf_2017}
Chelsea Finn and Sergey Levine.
\newblock {Deep Visual Foresight for Planning Robot Motion}.
\newblock In \emph{IEEE International Conference on Robotics and Automation
  (ICRA)}, 2017.

\bibitem[Gao et~al.(2016)Gao, Chang, and Demiris]{personalized_dressing_2016}
Yixing Gao, Hyung~Jin Chang, and Yiannis Demiris.
\newblock {Iterative Path Optimisation for Personalised Dressing Assistance
  using Vision and Force Information}.
\newblock In \emph{IEEE/RSJ International Conference on Intelligent Robots and
  Systems (IROS)}, 2016.

\bibitem[Goodfellow et~al.(2016)Goodfellow, Bengio, and
  Courville]{Goodfellow-et-al-2016}
Ian Goodfellow, Yoshua Bengio, and Aaron Courville.
\newblock \emph{{Deep Learning}}.
\newblock MIT Press, 2016.
\newblock \url{http://www.deeplearningbook.org}.

\bibitem[Hewing et~al.(2018)Hewing, Liniger, and Zeilinger]{cautious-MPC}
Lukas Hewing, Alexander Liniger, and Melanie Zeilinger.
\newblock {Cautious NMPC with Gaussian Process Dynamics for Autonomous
  Miniature Race Cars}.
\newblock In \emph{European Controls Conference (ECC)}, 2018.

\bibitem[Jangir et~al.(2020)Jangir, Alenya, and Torras]{rishabh_2019}
Rishabh Jangir, Guillem Alenya, and Carme Torras.
\newblock {Dynamic Cloth Manipulation with Deep Reinforcement Learning}.
\newblock In \emph{IEEE International Conference on Robotics and Automation
  (ICRA)}, 2020.

\bibitem[Jia et~al.(2018)Jia, Hu, Pan, and Manocha]{jia_visual_feedback_2018}
Biao Jia, Zhe Hu, Jia Pan, and Dinesh Manocha.
\newblock {Manipulating Highly Deformable Materials Using a Visual Feedback
  Dictionary}.
\newblock In \emph{IEEE International Conference on Robotics and Automation
  (ICRA)}, 2018.

\bibitem[Jia et~al.(2019)Jia, Pan, Hu, Pan, and Manocha]{jia_cloth_manip_2019}
Biao Jia, Zherong Pan, Zhe Hu, Jia Pan, and Dinesh Manocha.
\newblock {Cloth Manipulation Using Random-Forest-Based Imitation Learning}.
\newblock In \emph{IEEE International Conference on Robotics and Automation
  (ICRA)}, 2019.

\bibitem[Kazanzides et~al.(2014)Kazanzides, Chen, Deguet, Fischer, Taylor, and
  DiMaio]{dvrk2014}
P~Kazanzides, Z~Chen, A~Deguet, G~Fischer, R~Taylor, and S.~DiMaio.
\newblock {An Open-Source Research Kit for the da Vinci Surgical System}.
\newblock In \emph{IEEE International Conference on Robotics and Automation
  (ICRA)}, 2014.

\bibitem[Kingma and Ba(2015)]{adam2015}
Diederik~P. Kingma and Jimmy Ba.
\newblock {Adam: A Method for Stochastic Optimization}.
\newblock In \emph{International Conference on Learning Representations
  (ICLR)}, 2015.

\bibitem[Kita et~al.(2009{\natexlab{a}})Kita, Ueshiba, Neo, and
  Kita]{kita_2009_icra}
Yasuyo Kita, Toshio Ueshiba, Ee~Sian Neo, and Nobuyuki Kita.
\newblock {Clothes State Recognition Using 3D Observed Data}.
\newblock In \emph{IEEE International Conference on Robotics and Automation
  (ICRA)}, 2009{\natexlab{a}}.

\bibitem[Kita et~al.(2009{\natexlab{b}})Kita, Ueshiba, Neo, and
  Kita]{kita_2009_iros}
Yasuyo Kita, Toshio Ueshiba, Ee~Sian Neo, and Nobuyuki Kita.
\newblock {A Method For Handling a Specific Part of Clothing by Dual Arms}.
\newblock In \emph{IEEE/RSJ International Conference on Intelligent Robots and
  Systems (IROS)}, 2009{\natexlab{b}}.

\bibitem[Kocijan et~al.(2004)Kocijan, Murray-Smith, Rasmussen, and
  Girard]{GP-MPC}
Jus Kocijan, Roderick Murray-Smith, C.E. Rasmussen, and A.~Girard.
\newblock {Gaussian Process Model Based Predictive Control}.
\newblock In \emph{American Control Conference (ACC)}, 2004.

\bibitem[Li et~al.(2015)Li, Yue, Grinspun, and Allen]{folding_iros_2015}
Yinxiao Li, Yonghao Yue, Danfei Xu~Eitan Grinspun, and Peter~K. Allen.
\newblock {Folding Deformable Objects using Predictive Simulation and
  Trajectory Optimization}.
\newblock In \emph{IEEE/RSJ International Conference on Intelligent Robots and
  Systems (IROS)}, 2015.

\bibitem[Li et~al.(2016)Li, Hu, Xu, Yue, Grinspun, and Allen]{ironing_2016}
Yinxiao Li, Xiuhan Hu, Danfei Xu, Yonghao Yue, Eitan Grinspun, and Peter~K.
  Allen.
\newblock {Multi-Sensor Surface Analysis for Robotic Ironing}.
\newblock In \emph{IEEE International Conference on Robotics and Automation
  (ICRA)}, 2016.

\bibitem[Lillicrap et~al.(2016)Lillicrap, Hunt, Pritzel, Heess, Erez, Tassa,
  Silver, and Wierstra]{ddpg2016}
Timothy~P. Lillicrap, Jonathan~J. Hunt, Alexander Pritzel, Nicolas Heess, Tom
  Erez, Yuval Tassa, David Silver, and Daan Wierstra.
\newblock {Continuous Control with Deep Reinforcement Learning}.
\newblock In \emph{International Conference on Learning Representations
  (ICLR)}, 2016.

\bibitem[Maitin-Shepard et~al.(2010)Maitin-Shepard, Cusumano-Towner, Lei, and
  Abbeel]{maitin2010cloth}
Jeremy Maitin-Shepard, Marco Cusumano-Towner, Jinna Lei, and Pieter Abbeel.
\newblock {Cloth Grasp Point Detection Based on Multiple-View Geometric Cues
  with Application to Robotic Towel Folding}.
\newblock In \emph{IEEE International Conference on Robotics and Automation
  (ICRA)}, 2010.

\bibitem[Mann and Whitney(1947)]{mannwhitney}
Henry Mann and Donald Whitney.
\newblock {On a Test of Whether One of Two Random Variables is Stochastically
  Larger than the Other}.
\newblock \emph{Annals of Mathematical Statistics}, 1947.

\bibitem[Matas et~al.(2018)Matas, James, and Davison]{sim2real_deform_2018}
Jan Matas, Stephen James, and Andrew~J. Davison.
\newblock {Sim-to-Real Reinforcement Learning for Deformable Object
  Manipulation}.
\newblock \emph{Conference on Robot Learning (CoRL)}, 2018.

\bibitem[Miller et~al.(2012)Miller, van~den Berg, Fritz, Darrell, Goldberg, and
  Abbeel]{laundry2012}
Stephen Miller, Jur van~den Berg, Mario Fritz, Trevor Darrell, Ken Goldberg,
  and Pieter Abbeel.
\newblock {A Geometric Approach to Robotic Laundry Folding}.
\newblock In \emph{International Journal of Robotics Research (IJRR)}, 2012.

\bibitem[Nair et~al.(2018)Nair, McGrew, Andrychowicz, Zaremba, and
  Abbeel]{overcoming_exploration}
Ashvin Nair, Bob McGrew, Marcin Andrychowicz, Wojciech Zaremba, and Pieter
  Abbeel.
\newblock {Overcoming Exploration in Reinforcement Learning with
  Demonstrations}.
\newblock In \emph{IEEE International Conference on Robotics and Automation
  (ICRA)}, 2018.

\bibitem[Nair et~al.(2020)Nair, Babaeizadeh, Finn, Levine, and
  Kumar]{time_reversal}
Suraj Nair, Mohammad Babaeizadeh, Chelsea Finn, Sergey Levine, and Vikash
  Kumar.
\newblock {Time Reversal as Self-Supervision}.
\newblock In \emph{IEEE International Conference on Robotics and Automation
  (ICRA)}, 2020.

\bibitem[Narain et~al.(2012)Narain, Samii, and O'Brien]{arcsim2012}
Rahul Narain, Armin Samii, and James~F. O'Brien.
\newblock {Adaptive Anisotropic Remeshing for Cloth Simulation}.
\newblock In \emph{ACM SIGGRAPH Asia}, 2012.

\bibitem[OpenAI et~al.(2018)OpenAI, Andrychowicz, Baker, Chociej, Jozefowicz,
  McGrew, Pachocki, Petron, Plappert, Powell, Ray, Schneider, Sidor, Tobin,
  Welinder, Weng, and Zaremba]{openai-dactyl}
OpenAI, Marcin Andrychowicz, Bowen Baker, Maciek Chociej, Rafal Jozefowicz, Bob
  McGrew, Jakub Pachocki, Arthur Petron, Matthias Plappert, Glenn Powell, Alex
  Ray, Jonas Schneider, Szymon Sidor, Josh Tobin, Peter Welinder, Lilian Weng,
  and Wojciech Zaremba.
\newblock {Learning Dexterous In-Hand Manipulation}.
\newblock \emph{arXiv:1808.00177}, 2018.

\bibitem[Osawa et~al.(2007)Osawa, Seki, and Kamiya]{osawa_2007}
Fumiaki Osawa, Hiroaki Seki, and Yoshitsugu Kamiya.
\newblock {Unfolding of Massive Laundry and Classification Types by Dual
  Manipulator}.
\newblock \emph{Journal of Advanced Computational Intelligence and Intelligent
  Informatics}, 11\penalty0 (5), 2007.

\bibitem[Pomerleau(1991)]{Pomerleau_behavior_cloning}
Dean~A. Pomerleau.
\newblock {Efficient Training of Artificial Neural Networks for Autonomous
  Navigation}.
\newblock \emph{Neural Comput.}, 3, 1991.

\bibitem[Provot(1995)]{provot_1996}
Xavier Provot.
\newblock {Deformation Constraints in a Mass-Spring Model to Describe Rigid
  Cloth Behavior}.
\newblock In \emph{Graphics Interface}, 1995.

\bibitem[Rosolia and Borrelli(2019)]{MPCRacing}
Ugo Rosolia and Francesco Borrelli.
\newblock {Learning how to Autonomously Race a Car: a Predictive Control
  Approach}.
\newblock In \emph{IEEE Transactions on Control Systems Technology}, 2019.

\bibitem[Ross et~al.(2011)Ross, Gordon, and Bagnell]{ross2011reduction}
Stephane Ross, Geoffrey~J Gordon, and J~Andrew Bagnell.
\newblock {A Reduction of Imitation Learning and Structured Prediction to
  No-Regret Online Learning}.
\newblock In \emph{International Conference on Artificial Intelligence and
  Statistics (AISTATS)}, 2011.

\bibitem[Rubinstein(1999)]{cem_1999}
Reuven Rubinstein.
\newblock {The Cross-Entropy Method for Combinatorial and Continuous
  Optimization}.
\newblock \emph{Methodology And Computing In Applied Probability}, 1999.

\bibitem[Sanchez et~al.(2018)Sanchez, Corrales, Bouzgarrou, and
  Mezouar]{manip_deformable_survey_2018}
Jose Sanchez, Juan-Antonio Corrales, Belhassen-Chedli Bouzgarrou, and Youcef
  Mezouar.
\newblock {Robotic Manipulation and Sensing of Deformable Objects in Domestic
  and Industrial Applications: a Survey}.
\newblock In \emph{International Journal of Robotics Research (IJRR)}, 2018.

\bibitem[Schrimpf and Wetterwald(2012)]{sewing_2012}
Johannes Schrimpf and Lars~Erik Wetterwald.
\newblock {Experiments Towards Automated Sewing With a Multi-Robot System}.
\newblock In \emph{IEEE International Conference on Robotics and Automation
  (ICRA)}, 2012.

\bibitem[Seita et~al.(2018)Seita, Krishnan, Fox, McKinley, Canny, and
  Goldberg]{seita_icra_2018}
Daniel Seita, Sanjay Krishnan, Roy Fox, Stephen McKinley, John Canny, and
  Kenneth Goldberg.
\newblock {Fast and Reliable Autonomous Surgical Debridement with Cable-Driven
  Robots Using a Two-Phase Calibration Procedure}.
\newblock In \emph{IEEE International Conference on Robotics and Automation
  (ICRA)}, 2018.

\bibitem[Seita et~al.(2019{\natexlab{a}})Seita, Ganapathi, Hoque, Hwang, Cen,
  Tanwani, Balakrishna, Thananjeyan, Ichnowski, Jamali, Yamane, Iba, Canny, and
  Goldberg]{seita_ryan}
Daniel Seita, Aditya Ganapathi, Ryan Hoque, Minho Hwang, Edward Cen, Ajay~Kumar
  Tanwani, Ashwin Balakrishna, Brijen Thananjeyan, Jeffrey Ichnowski, Nawid
  Jamali, Katsu Yamane, Soshi Iba, John Canny, and Ken Goldberg.
\newblock {Deep Imitation Learning of Sequential Fabric Smoothing Policies}.
\newblock \emph{arXiv:1910.04854}, 2019{\natexlab{a}}.

\bibitem[Seita et~al.(2019{\natexlab{b}})Seita, Jamali, Laskey, Berenstein,
  Tanwani, Baskaran, Iba, Canny, and Goldberg]{seita-bedmaking}
Daniel Seita, Nawid Jamali, Michael Laskey, Ron Berenstein, Ajay~Kumar Tanwani,
  Prakash Baskaran, Soshi Iba, John Canny, and Ken Goldberg.
\newblock {Deep Transfer Learning of Pick Points on Fabric for Robot
  Bed-Making}.
\newblock In \emph{International Symposium on Robotics Research (ISRR)},
  2019{\natexlab{b}}.

\bibitem[Shibata et~al.(2012)Shibata, Yoshimi, Mizukawa, and
  Ando]{shibata2012trajectory}
Syohei Shibata, Takashi Yoshimi, Makoto Mizukawa, and Yoshinobu Ando.
\newblock {A Trajectory Generation of Cloth Object Folding Motion Toward
  Realization of Housekeeping Robot}.
\newblock In \emph{International Conference on Ubiquitous Robots and Ambient
  Intelligence (URAI)}, 2012.

\bibitem[Shin et~al.(2019)Shin, Ferguson, Pedram, Ma, Dutson, and
  Rosen]{rosen_icra_tissues_2019}
Changyeob Shin, Peter~Walker Ferguson, Sahba~Aghajani Pedram, Ji~Ma, Erik~P.
  Dutson, and Jacob Rosen.
\newblock {Autonomous Tissue Manipulation via Surgical Robot Using Learning
  Based Model Predictive Control}.
\newblock In \emph{IEEE International Conference on Robotics and Automation
  (ICRA)}, 2019.

\bibitem[Sun et~al.(2014)Sun, Aragon-Camarasa, Cockshott, Rogers, and
  Siebert]{heuristic_wrinkles_2014}
Li~Sun, Gerarado Aragon-Camarasa, Paul Cockshott, Simon Rogers, and J.~Paul
  Siebert.
\newblock {A Heuristic-Based Approach for Flattening Wrinkled Clothes}.
\newblock \emph{Towards Autonomous Robotic Systems. TAROS 2013. Lecture Notes
  in Computer Science, vol 8069}, 2014.

\bibitem[Sun et~al.(2015)Sun, Aragon-Camarasa, Rogers, and
  Siebert]{cloth_icra_2015}
Li~Sun, Gerardo Aragon-Camarasa, Simon Rogers, and J.~Paul Siebert.
\newblock {Accurate Garment Surface Analysis using an Active Stereo Robot Head
  with Application to Dual-Arm Flattening}.
\newblock In \emph{IEEE International Conference on Robotics and Automation
  (ICRA)}, 2015.

\bibitem[Thananjeyan et~al.(2017)Thananjeyan, Garg, Krishnan, Chen, Miller, and
  Goldberg]{thananjeyan2017multilateral}
Brijen Thananjeyan, Animesh Garg, Sanjay Krishnan, Carolyn Chen, Lauren Miller,
  and Ken Goldberg.
\newblock {Multilateral Surgical Pattern Cutting in 2D Orthotropic Gauze with
  Deep Reinforcement Learning Policies for Tensioning}.
\newblock In \emph{IEEE International Conference on Robotics and Automation
  (ICRA)}, 2017.

\bibitem[Thananjeyan et~al.(2020)Thananjeyan, Balakrishna, Rosolia, Li,
  McAllister, Gonzalez, Levine, Borrelli, and Goldberg]{thananjeyan2019safety}
Brijen Thananjeyan, Ashwin Balakrishna, Ugo Rosolia, Felix Li, Rowan
  McAllister, Joseph~E Gonzalez, Sergey Levine, Francesco Borrelli, and Ken
  Goldberg.
\newblock {Safety Augmented Value Estimation from Demonstrations (SAVED): Safe
  Deep Model-Based RL for Sparse Cost Robotic Tasks}.
\newblock \emph{IEEE International Conference on Robotics and Automation
  (ICRA)}, 2020.

\bibitem[Tobin et~al.(2017)Tobin, Fong, Ray, Schneider, Zaremba, and
  Abbeel]{domain_randomization}
Josh Tobin, Rachel Fong, Alex Ray, Jonas Schneider, Wojciech Zaremba, and
  Pieter Abbeel.
\newblock {Domain Randomization for Transferring Deep Neural Networks from
  Simulation to the Real World}.
\newblock In \emph{IEEE/RSJ International Conference on Intelligent Robots and
  Systems (IROS)}, 2017.

\bibitem[Todorov et~al.(2012)Todorov, Erez, and Tassa]{mujoco}
Emanuel Todorov, Tom Erez, and Yuval Tassa.
\newblock {MuJoCo: A Physics Engine for Model-Based Control}.
\newblock In \emph{IEEE/RSJ International Conference on Intelligent Robots and
  Systems (IROS)}, 2012.

\bibitem[Torgerson and Paul(1987)]{Torgerson1987VisionGR}
Eric Torgerson and Fanget Paul.
\newblock {Vision Guided Robotic Fabric Manipulation for Apparel
  Manufacturing}.
\newblock In \emph{IEEE International Conference on Robotics and Automation
  (ICRA)}, 1987.

\bibitem[Vaswani et~al.(2018)Vaswani, Bengio, Brevdo, Chollet, Gomez, Gouws,
  Jones, Kaiser, Kalchbrenner, Parmar, Sepassi, Shazeer, and
  Uszkoreit]{tensor2tensor}
Ashish Vaswani, Samy Bengio, Eugene Brevdo, Francois Chollet, Aidan~N. Gomez,
  Stephan Gouws, Llion Jones, \L{}ukasz Kaiser, Nal Kalchbrenner, Niki Parmar,
  Ryan Sepassi, Noam Shazeer, and Jakob Uszkoreit.
\newblock Tensor2tensor for neural machine translation.
\newblock \emph{CoRR}, abs/1803.07416, 2018.
\newblock URL \url{http://arxiv.org/abs/1803.07416}.

\bibitem[Vecerik et~al.(2017)Vecerik, Hester, Scholz, Wang, Pietquin, Piot,
  Heess, Rothörl, Lampe, and Riedmiller]{ddpgfd}
Mel Vecerik, Todd Hester, Jonathan Scholz, Fumin Wang, Olivier Pietquin, Bilal
  Piot, Nicolas Heess, Thomas Rothörl, Thomas Lampe, and Martin Riedmiller.
\newblock {Leveraging Demonstrations for Deep Reinforcement Learning on
  Robotics Problems with Sparse Rewards}.
\newblock \emph{arXiv:1707.08817}, 2017.

\bibitem[Verlet(1967)]{verlet_1967}
Loup Verlet.
\newblock {Computer Experiments on Classical Fluids: I. Thermodynamical
  Properties of Lennard-Jones Molecules}.
\newblock \emph{Physics Review}, 159\penalty0 (98), 1967.

\bibitem[Wang et~al.(2004)Wang, Bovik, Sheikh, and Simoncelli]{ssim_2004}
Zhou Wang, A.~C. Bovik, H.~R. Sheikh, and E.~P. Simoncelli.
\newblock {Image Quality Assessment: From Error Visibility to Structural
  Similarity}.
\newblock \emph{Trans. Img. Proc.}, April 2004.

\bibitem[Willimon et~al.(2011)Willimon, Birchfield, and
  Walker]{willimon_unfolding_laundry_2011}
Bryan Willimon, Stan Birchfield, and Ian Walker.
\newblock {Model for Unfolding Laundry using Interactive Perception}.
\newblock In \emph{IEEE/RSJ International Conference on Intelligent Robots and
  Systems (IROS)}, 2011.

\bibitem[Wu et~al.(2019)Wu, Yan, Kurutach, Pinto, and Abbeel]{lerrel}
Yilin Wu, Wilson Yan, Thanard Kurutach, Lerrel Pinto, and Pieter Abbeel.
\newblock {Learning to Manipulate Deformable Objects without Demonstrations}.
\newblock \emph{arXiv:1910.13439}, 2019.

\bibitem[Yang et~al.(2017)Yang, Sasaki, Suzuki, Kase, Sugano, and
  Ogata]{folding_2017}
Pin-Chu Yang, Kazuma Sasaki, Kanata Suzuki, Kei Kase, Shigeki Sugano, and
  Tetsuya Ogata.
\newblock {Repeatable Folding Task by Humanoid Robot Worker Using Deep
  Learning}.
\newblock In \emph{IEEE Robotics and Automation Letters (RA-L)}, 2017.

\end{thebibliography}

\clearpage
\normalsize
\appendices

\section{Fabric Simulators}\label{app:simulators}

In this work, we use the fabric simulator from~\citet{seita_ryan}. This simulator possesses an ideal balance between ease of code implementation, speed, and accuracy, and was able to lead to reasonable smoothing policies in prior work. We considered using simulators from ARCSim~\cite{arcsim2012}, MuJoCo~\cite{mujoco}, PyBullet~\cite{coumans2019}, or Blender~\cite{blender},  but did not use them for several reasons. 

High-fidelity simulators, such as ARCSim, take too long to simulate to get sufficient data for training visual dynamics models. We attempted to simulate rudimentary grasping behavior in ARCSim, but it proved difficult because ARCSim does not represent fabric as a fixed grid of vertices, which meant we could not simulate grasping by ``pinning'' or ``fixing'' certain vertices.

The MuJoCo fabric simulator was only recently released in October 2018, and besides concurrent work from~\citet{lerrel}, there are no existing environments that combine fabrics with simulated robot grasps. We investigated and used the open-source code from~\citet{lerrel}, but found that MuJoCo did not accurately simulate fabric-fabric collisions well.

The PyBullet simulator code from~\citet{sim2real_deform_2018} showed relatively successful fabric simulation, but it was difficult for us to adapt the author's code to the proposed work, which made significant changes to the off-the-shelf PyBullet code.

Blender includes a new fabric simulator, with substantial improvements after 2017 for more realistic shearing and tensioning. These changes, however, are only supported in Blender 2.8, not Blender 2.79, and we used 2.79 because Blender 2.8 does not allow background processes to run on headless servers, which prevented us from running mass data collection.

Most of these fabric simulators were only recently developed, and some developed concurrently with this work. We will further investigate the feasibility of using these simulators.

\section{Details of Learning-Based Methods}\label{app:implementation}

\begin{figure}[t]
\center
\includegraphics[width=0.48\textwidth]{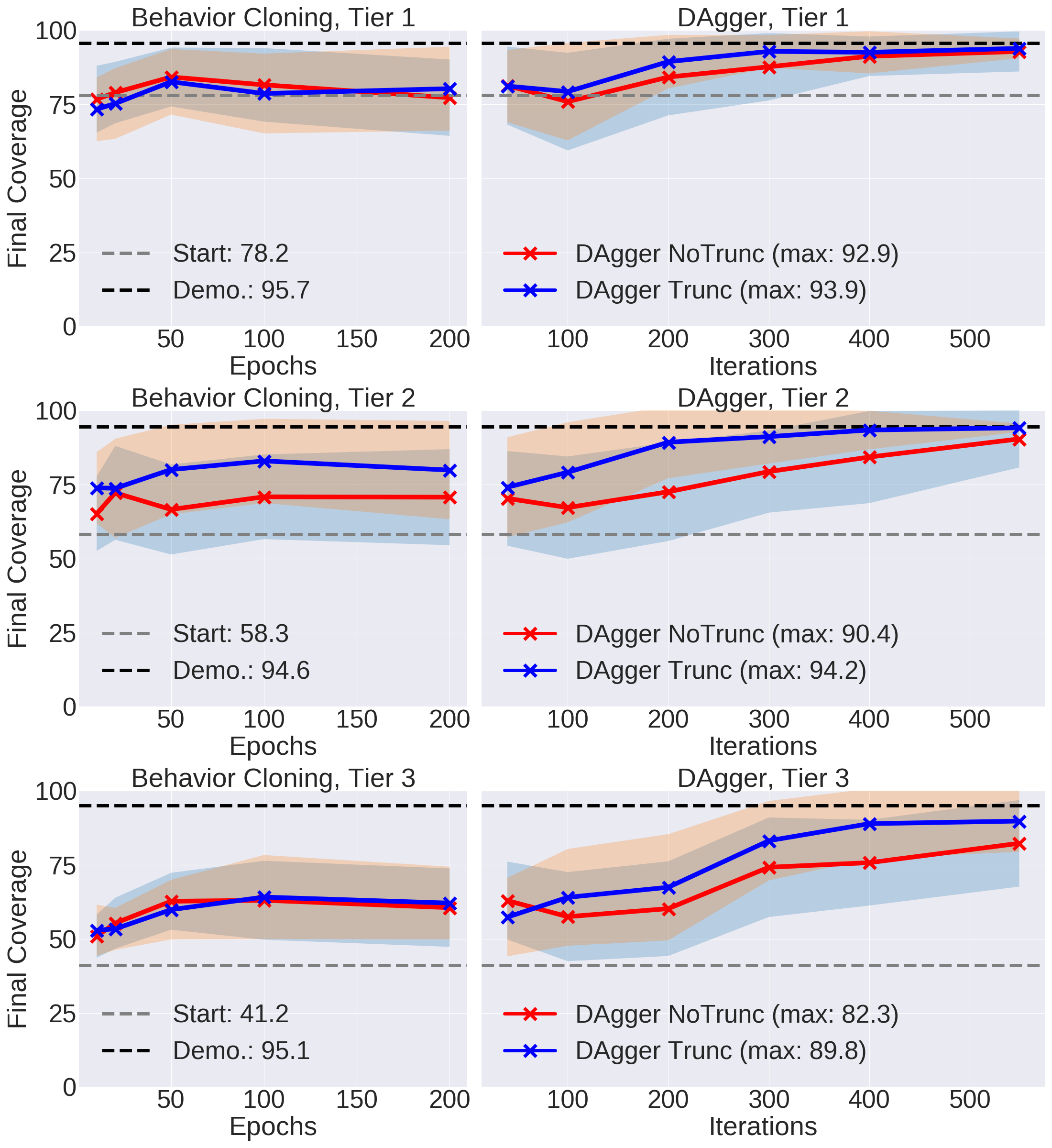}
\caption{
\small
Average coverage over 50 simulated test-time episodes at checkpoints (marked ``X'') during the behavior cloning and DAgger phases. For each setting of no action truncation and action truncation, we deploy a single DAgger policy deployed on all tiers. Using dashed lines, we annotate the average starting coverage and the corner pulling demonstrator's average final coverage.
}
\label{fig:10-0-1}
\end{figure}

We describe implementation and training details of the three learning-based methods tested: imitation learning, model-free reinforcement learning, and model-based \vsf. The other baselines --- random, highest point, and wrinkles --- are borrowed unmodified from prior open-source code~\cite{seita_ryan}. To ensure that comparisons are reasonably fair among the methods, we keep hyperparameters as similar as possible.

\subsection{Imitation Learning Baseline: DAgger}

DAgger~\cite{ross2011reduction} is implemented directly from the open source DAgger code in~\citet{seita_ryan}. This was originally based on the open-source OpenAI baselines~\cite{baselines} library for parallel environment support to overcome the time bottleneck of fabric simulation.

We ran the corner pulling demonstrator for 2,000 trajectories, resulting in 6,697 image-action pairs $(\bo_t, \ba_t')$, where the notation $\ba_t'$ indicates the action is labeled and comes from the demonstrator. Each trajectory was randomly drawn from one of the three tiers in the simulator with equal probability. We then perform a behavior cloning~\cite{Pomerleau_behavior_cloning} ``pre-training'' period for 200 epochs over this offline data, which does not require environment interaction.

After behavior cloning, each DAgger iteration rolls out 20 parallel environments for 10 steps each (hence, 200 total new samples) which are labeled by the corner pulling policy, the same policy that created the offline data and uses underlying state information. These are inserted into a replay buffer of image-action samples, where all samples have actions labeled by the demonstrator. The replay buffer size is 50,000, but the original demonstrator data of size 6,697 is never removed from it. After environment stepping, we draw 240 minibatches of size 128 each for training and use Adam~\cite{adam2015} for optimization. The process repeats with the agent rolling out its updated policy. We run DAgger for 110,000 steps across all environments (hence, 5,500 steps per parallel environment) to make the number of actions consumed to be roughly the same as the number of actions used to train the video prediction model. This is significantly more than the 50,000 DAgger training steps in prior work~\cite{seita_ryan}. Table~\ref{tab:dagger-hyperparams} contains additional hyperparameters.

The actor (i.e., policy) neural network for DAgger uses a design based on~\citet{seita_ryan} and~\citet{sim2real_deform_2018}. The input to the policy are RGBD images of size $(56 \times 56 \times 4)$, where the four channels are formed from stacking an RGB and a single-channel depth image. The policy processes the input through four convolutional layers that have 32 filters with size $3\times 3$, and then uses four fully connected layers with 256 nodes each. The parameters of the network, ignoring biases for simplicity, are listed as follows:

\footnotesize
\begin{verbatim}
actor/convnet/c1/w   1152 params (3, 3, 4, 32)
actor/convnet/c2/w   9216 params (3, 3, 32, 32)
actor/convnet/c3/w   9216 params (3, 3, 32, 32)
actor/convnet/c4/w   9216 params (3, 3, 32, 32)
actor/fcnet/fc1      663552 params (2592, 256)
actor/fcnet/fc2      65536 params (256, 256)
actor/fcnet/fc3      65536 params (256, 256)
actor/fcnet/fc4      1024 params (256, 4)
Total model parameters: 0.83 million
\end{verbatim}
\normalsize

The result from the actor policy is a 4D vector representing the action choice $\ba_t \in \mathbb{R}^4$ at each time step $t$. The last layer is a hyperbolic tangent which makes each of the four components of $\ba_t$ within $[-1,1]$. During action truncation, we further limit the two components of $\ba_t$ corresponding to the deltas into $[-0.4, 0.4]$.

A set of graphs representing learning progress for DAgger is shown in Figure~\ref{fig:10-0-1}, where for each marked snapshot, we roll it out in the environment for 50 episodes and measure final coverage. Results suggest the single DAgger policy, when trained with 110,000 total steps on RGBD images, performs well on all three tiers with performance nearly matching the 95-96\% coverage of the demonstrator.

We trained two variants of DAgger, one with and one without the action truncation to $[-0.4, 0.4]$ for the two deltas $\Delta x$ and $\Delta y$. The model trained on truncated actions outperforms the alternative setting. Furthermore, it is the setting we use in \vsf, hence we use it for physical robot experiments. We choose the final snapshot as it has the highest test-time performance, and we use it as the policy for simulated and real benchmarks in the main part of the paper.

\begin{table}[t]
\caption{
\small
DAgger hyperparameters. 
}
\centering
\begin{tabular}{l r}
\textbf{Hyperparameter} & \textbf{Value} \\  \hline
Parallel environments & 20 \\
Steps per env, between gradient updates & 10 \\
Gradient updates after parallel steps & 240 \\
Minibatch size            & 128 \\
Discount factor $\gamma$ & 0.99 \\
Demonstrator (offline) samples & 6697 \\
Policy learning rate   & 1e-4 \\
Policy $L_2$ regularization parameter  & 1e-5 \\
Behavior Cloning epochs & 200 \\
DAgger steps after Behavior Cloning    & 110000 \\
\end{tabular}
\label{tab:dagger-hyperparams}
\end{table}

\subsection{Model-Free Reinforcement Learning Baseline: DDPG}

\begin{figure}[t]
\center
\includegraphics[width=0.48\textwidth]{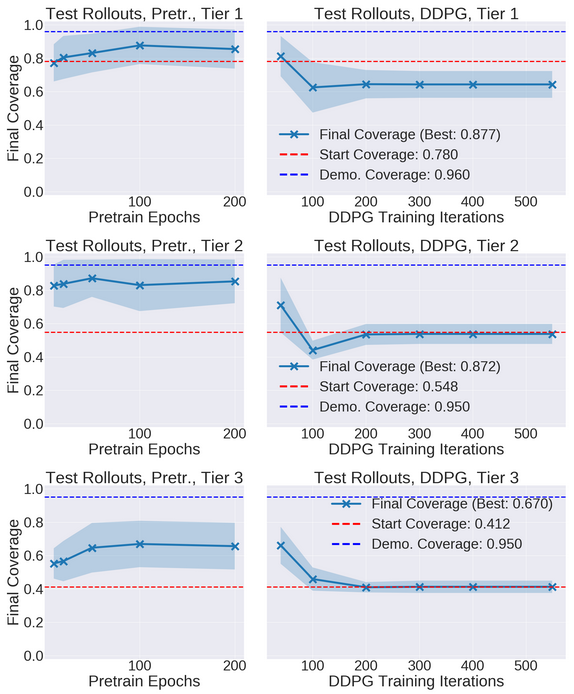}
\caption{
\small
Average coverage over 50 simulated test-time episodes at checkpoints (marked ``X'') during the pre-training DDPG phase, and the DDPG phase with agent exploration. This is presented in a similar manner as in Figure~\ref{fig:10-0-1} for DAgger. Results suggest that DDPG has difficulty in training a policy that can achieve high coverage.
}
\label{fig:10-1-0}
\end{figure}

\begin{table}[t]
\caption{
\small
DDPG hyperparameters. 
}
\centering
\begin{tabular}{l r}
\textbf{Hyperparameter} & \textbf{Value} \\  \hline
Parallel environments & 20 \\
Steps per env, between gradient updates & 10 \\
Gradient updates after parallel steps & 240 \\
Minibatch size            & 128 \\
Discount factor $\gamma$ & 0.99 \\
Demonstrator (offline) samples & 6697 \\
Actor learning rate   & 1e-4 \\
Actor $L_2$ regularization parameter  & 1e-5 \\
Critic learning rate   & 1e-3 \\
Critic $L_2$ regularization parameter  & 1e-5 \\
Pre-training epochs  & 200 \\
DDPG steps after pre-training & 110000 \\
\end{tabular}
\label{tab:ddpg-hyperparams}
\end{table}

To provide a second competitive baseline, we apply  model-free reinforcement learning. Specifically, we use a variant of Deep Deterministic Policy Gradients (DDPG)~\cite{ddpg2016} with several improvements as proposed in the research literature.  Briefly, DDPG is a deep reinforcement learning algorithm which trains parameterized actor and critic models, each of which are normally neural networks. The actor is the policy, and the critic is a value function.

First, as with DAgger, we use demonstrations~\cite{ddpgfd} to improve the performance of the learned policy. We use the same demonstrator data of 6,697 samples from DAgger, except this time each sample is a tuple of $(\bo_t, \ba_t', r_t, \bo_{t+1})$, including a scalar reward $r_t$ (to be described) and a successor state $\bo_{t+1}$. This data is added to the replay buffer and never removed.

We use a pre-training phase (of 200 epochs) to initialize the actor and critic. We also apply $L_2$ regularization for both the actor and critic networks. In addition, we use the Q-filter from~\citet{overcoming_exploration} which may help the actor learn better actions than the demonstrator provides, perhaps for cases when naive corner pulling might not be ideal.

For a fairer comparison, the actor network for DDPG uses the same architecture as the actor for DAgger. The critic has a similar architecture as the actor, with the only change that the action input $\ba_t$ is inserted and concatenated with the features of the image $\bo_t$ after the four convolutional layers, and before the fully connected portion. As with the imitation learning baseline, the inputs are RGBD images of size $(56\times 56 \times 4)$.

We design a dense reward to encourage the agent to achieve high coverage. At each time, the agent gets reward based on:

\begin{itemize}
    \item A small negative living reward of -0.05
    \item A small negative reward of -0.05 for failing to grasp any point on the fabric (i.e., a wasted grasp attempt).
    \item A delta in coverage based on the change in coverage from the current state and the prior state.
    \item A +5 bonus for triggering 92\% coverage.
    \item A -5 penalty for triggering an out-of-bounds condition, where the fabric significantly exceeds the boundaries of the underlying fabric plane.
\end{itemize}

We designed the reward function by informal tuning and borrowing ideas from the reward in~\citet{openai-dactyl}, which used a delta in joint angles and a similar bonus for moving a block towards a target, or a penalty for dropping it. Intuitively, an agent may learn to take a slightly counter-productive action which would decrease coverage (and thus the delta reward component is negative), but which may enable an easier subsequent action to trigger a high bonus. This reward design is only suited for smoothing. As with the imitation learning baseline, the model-free DDPG baseline is not designed for non-smoothing tasks.

Figure~\ref{fig:10-1-0} suggests that the pre-training phase, where the actor and critic are trained on the demonstrator data, helps increase coverage. The DDPG portion of training, however, results in performance collapse to achieving no net coverage. Upon further inspection, this is because the actions collapsed to having no ``deltas,'' so the robot reduces to picking up but then immediately releasing the fabric. Due to the weak performance of DDPG, we do not benchmark the policy on the physical robot.

\subsection{VisuoSpatial Foresight}\label{app:visualmpc}

\begin{table}[t]
\caption{
\small
Visual MPC hyperparameters. 
}
\centering
\begin{tabular}{l r}
\textbf{Hyperparameter} & \textbf{Value} \\  \hline
Number of CEM iterations & 10 \\
CEM population size & 2000 \\
CEM $\alpha$ & 0.1 \\
CEM planning horizon & 5 \\
CEM initial mean $\mu$ & (0, 0, 0, 0) \\
CEM initial variance $\Sigma$ & (0.67, 0.67, 0.24, 0.24) \\
\end{tabular}
\label{tab:vismpc-hyperparams}
\end{table}

\begin{figure}[t]
\center
\includegraphics[width=0.48\textwidth]{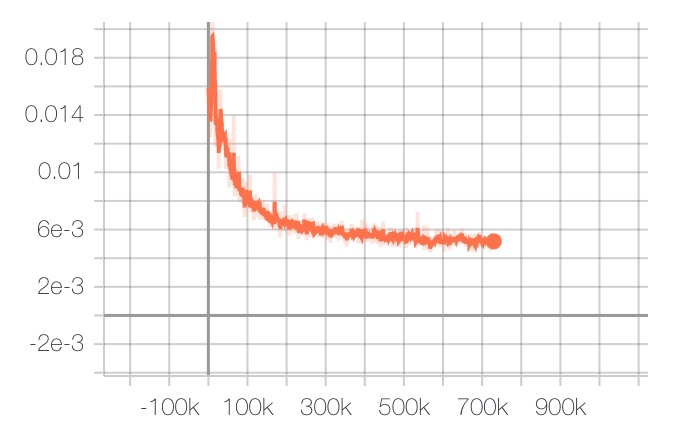}
\caption{
\small
VSF dynamics model loss curve.
}
\label{fig:losscurve}
\end{figure}

The main technique considered in this paper is VisuoSpatial Foresight (VSF), an extension of Visual Foresight~\cite{visual_foresight_2018}. It consists of a training phase followed by a planning phase. The training phase is as described in Section~\ref{ssec:visual}: we collect 7,003 random episodes with 15 actions and 16 $56\times 56 \times 4$ RGBD observations each. We use the SV2P \cite{sv2p} implementation in \cite{tensor2tensor}. We set the number of input channels to 4 for RGBD data and predict $H = 7$ output frames from $m = 3$ context frames. During planning, we predict $H = 5$ output frames from $m = 1$ context frame. See Figure~\ref{fig:losscurve} for the loss curve.

For the planning phase, we tuned the hyperparameters in Table~\ref{tab:vismpc-hyperparams}. The variance reported in the table is the diagonal covariance used for folding and double folding. We found that for smoothing, a lower variance (0.25, 0.25, 0.04, 0.04) results in better performance, although it does push the policy toward shorter actions. For the pixel distance cost function, we remove the 7 pixels on each side of the image to get rid of the impact of the dark border, i.e. we turn a $56 \times 56$ frame into a $42\times 42$ frame.

\section{Supplementary Smoothing Results}\label{app:results}

\subsection{Statistical Significance Tests}

We run the Mann-Whitney U test~\cite{mannwhitney} on the coverage and number of action results reported in Table~\ref{tab:analytic} for VSF against all baselines other than Imitation Learning, which we wish to perform similarly to. See Table~\ref{tab:stat} for computed $p$-values. We conclude that we can confidently reject the null hypothesis that the values are drawn from the same distribution for all metrics except Tier 2 coverage for Wrinkle and the Tier 1 and Tier 3 number of actions for DDPG ($p$ $<$ 0.02). Note that Tier 3 results are most informative, as it is the most difficult tier.

\begin{table}[t]
\caption{
\small
Mann-Whitney Test $p$-values for coverage and number of actions on VSF compared with Random, Highest, Wrinkle and DDPG baselines across all tiers of difficulty.
}
\centering
\begin{tabular}{c c c c}
\textbf{Tier} & \textbf{Policy} & \textbf{Coverage $p$-value} & \textbf{Actions $p$-value} \\  \hline
1 & Random & 0.0000 & 0.0000 \\
1 & Highest & 0.0000 & 0.0002 \\
1 & Wrinkle & 0.0040 & 0.0015 \\
1 & DDPG & 0.0044 & 0.3670 \\
2 & Random & 0.0000 & 0.0000 \\
2 & Highest & 0.0000 & 0.0000 \\
2 & Wrinkle & 0.2323 & 0.0091 \\
2 & DDPG & 0.0000 & 0.0000 \\
3 & Random & 0.0000 & 0.0000 \\
3 & Highest & 0.0000 & 0.0000 \\
3 & Wrinkle & 0.0030 & 0.0199 \\
3 & DDPG & 0.0000 & 0.0674 \\
\end{tabular}
\label{tab:stat}
\end{table}

\subsection{Domain Randomization Ablation}

We run 50 simulated smoothing episodes per tier with a policy trained \textit{without} domain randomization and compare with the 200 episodes from Table~\ref{tab:analytic}. In the episodes without domain randomization, we keep fabric color, camera angle, background plane shading, and brightness constant at training and testing time. In the episodes with domain randomization, we randomize these parameters in the training data and test in the same setting as the experiments without domain randomization, which can be interpreted as a random initialization of the domain randomized parameters. In particular, we vary the following:

\begin{itemize}
    \item Fabric color RGB values uniformly between (0, 0, 128) and (115, 179, 255), centered around blue.
    \item Background plane color RGB values uniformly between (102, 102, 102) and (153, 153, 153).
    \item RGB gamma correction with gamma uniformly between 0.7 and 1.3.
    \item A fixed amount to subtract from the depth image between 40 and 50 to simulate changing the height of the depth camera.
    \item Camera position ($x, y, z$) as $(0.5+\delta_1, 0.5+\delta_2, 1.45+\delta_3)$ meters, where each $\delta_i$ is sampled from $\mathcal{N}(0, 0.04)$.
    \item Camera rotation with Euler angles sampled from $\mathcal{N}(0, 90^{\degree} )$.
    \item Random noise at each pixel uniformly between -15 and 15.
\end{itemize}

From the results in Table~\ref{tab:dr}, we find that the final coverage values are very similar whether or not we use domain randomization on training data, and we conclude our domain randomization techniques do not have an adverse effect on performance in simulation.

To analyze robustness of the policy to variation in the randomized parameters, we also evaluate the former two policies (trained with and without domain randomization) with randomization in the test environment on Tier 3 starting states. Specifically, we change the color of the fabric in fixed increments from its non-randomized setting (RGB (25, 89, 217)) until performance starts to deteriorate. In Table~\ref{tab:dr2}, we observe that the domain randomized policy maintains high coverage within the training range (RGB (0, 0, 128) to (115, 179, 255)) while the policy without domain randomization suffers as soon as the fabric color is slightly altered.

\begin{table}[t]
\caption{
\small
Coverage and number of actions for simulated smoothing episodes with and without domain randomization on training data, where the domain randomized results are from Table~\ref{tab:analytic}.
}
\centering
\begin{tabular}{c c r r}
\textbf{Tier} & \textbf{Domain Randomized?} & \textbf{Coverage} & \textbf{Actions} \\ \hline
1 & Yes & 92.5 $\pm$ \:\:2.5 & 8.3 $\pm$ 4.7 \\
1 & No & 93.0 $\pm$ \:\:3.0 & 6.9 $\pm$ 4.1 \\
2 & Yes & 90.3 $\pm$ \:\:3.8 & 12.1 $\pm$ 3.4 \\ 
2 & No & 91.2 $\pm$ \:\:9.2 & 8.7 $\pm$ 3.6 \\
3 & Yes & 89.3 $\pm$ \:\:5.9 & 13.1 $\pm$ 2.9 \\
3 & No & 85.1 $\pm$ 12.8 & 9.9 $\pm$ 3.9 \\
\end{tabular}
\label{tab:dr}
\end{table}

\begin{table}[t]
\caption{
\small
Coverage and number of actions for Tier 3 simulated smoothing episodes with and without domain randomization on training data, where we vary fabric color in fixed increments. (26, 89, 217) is the default blue color and (128, 191, 115) is slightly outside the domain randomization range. Values for the default setting are repeated from Table~\ref{tab:dr} and all other data points are averaged over 20 episodes.
}
\centering
\begin{tabular}{c c r r}
\textbf{RGB Values} & \textbf{DR?} & \textbf{Coverage} & \textbf{Actions} \\ \hline
(26, 89, 217) & Yes & 89.3 $\pm$ \:\:5.9 & 13.1 $\pm$ 2.9 \\
(51, 115, 191) & Yes & 89.3 $\pm$ 10.3 & 11.7 $\pm$ 3.5 \\
(77, 140, 166) & Yes & 91.4 $\pm$ \:\:3.1 & 11.7 $\pm$ 3.1 \\ 
(102, 165, 140) & Yes & 85.6 $\pm$ 10.1 & 13.2 $\pm$ 2.7 \\
(128, 191, 115) & Yes & 54.7 $\pm$ \:\:6.5 & 10.3 $\pm$ 4.0 \\
(26, 89, 217) & No & 85.1 $\pm$ 12.8 & 9.9 $\pm$ 3.9 \\
(51, 115, 191) & No & 60.7 $\pm$ 13.6 & 7.4 $\pm$ 2.4 \\
\end{tabular}
\label{tab:dr2}
\end{table}

\subsection{Action Magnitudes}\label{app:act-mag}

\begin{figure}[h]
\center
\includegraphics[width=0.45\textwidth]{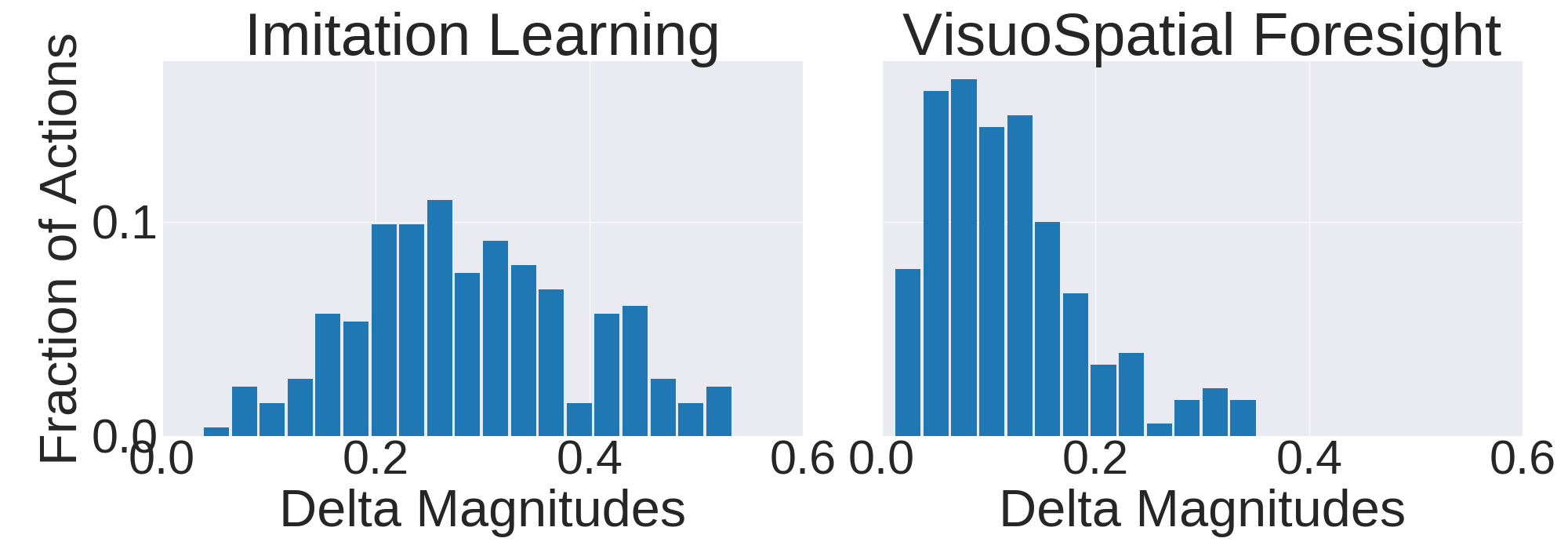}
\caption{
\small
Histograms showing the distribution of action magnitudes $\sqrt{(\Delta x)^2 + (\Delta y)^2}$ taken by the Imitation Learning and VisuoSpatial Foresight policies from the physical experiments reported in Table~\ref{tab:surgical}. The y-axis reports each bin as a fraction of the number of actions. The x-axis is consistent among both plots, showing that VSF takes actions with smaller deltas, likely due to the initialization of the mean and variance used to fit the conditional Gaussians for CEM.
}
\label{fig:action_deltas}
\end{figure}

Figure~\ref{fig:action_deltas} shows histograms of the action delta magnitudes, computed as $\sqrt{(\Delta x)^2+(\Delta y)^2}$, for the physical experiments reported in Section~\ref{ssec:physical-smoothing}. The histograms strongly suggest that the action magnitudes are smaller for VSF as compared to IL.

\end{document}